\documentclass[10pt,twocolumn,letterpaper]{article}
\usepackage[pagenumbers]{iccv} 
\usepackage[accsupp]{axessibility}
\usepackage{color}
\usepackage{soul}
\usepackage{amssymb}
\usepackage{pifont}
\usepackage{algorithm}
\usepackage{algorithmic}
\newcommand{\cmark}{\ding{51}}
\newcommand{\xmark}{\ding{55}}
\newcommand{\dataset}{CT-ScanGaze\xspace}
\newcommand{\method}{CT-Searcher\xspace}
\usepackage{multirow}
\usepackage{listings}
\usepackage[utf8]{inputenc}
\usepackage{xcolor}

\definecolor{blue_2}{RGB}{195, 222, 231}
\definecolor{orange_2}{RGB}{245, 188, 129}
\definecolor{green_2}{RGB}{67, 150, 85}

\definecolor{iccvblue}{rgb}{0.21,0.49,0.74}

\usepackage[pagebackref,breaklinks,colorlinks,allcolors=iccvblue]{hyperref}

\def\paperID{4282} 
\def\confName{ICCV}
\def\confYear{2025}

\title{CT-ScanGaze: A Dataset and Baselines for 3D Volumetric Scanpath Modeling}

\author{Trong Thang Pham$^{1}$, Akash Awasthi$^{2}$, Saba Khan$^{2}$, Esteban Duran Marti$^{1}$,\\ 
Tien-Phat Nguyen$^{3}$, Khoa Vo$^{1}$, Minh Tran$^{1}$, Ngoc Son Nguyen$^{4}$, Cuong Tran Van$^{4}$, Yuki Ikebe$^{1}$, \\
Anh Totti Nguyen$^{5}$, Anh Nguyen$^{6}$, Zhigang Deng$^{2}$, Carol C. Wu$^{7}$, Hien Nguyen$^{2}$, and Ngan Le$^{1}$ \\ \\ 
{\normalsize $^{1}$University of Arkansas,
$^{2}$University of Houston, $^{3}$University of Science VNU-HCM, }\\ 
{\normalsize $^{4}$FPT Software, $^{5}$Auburn University,
$^{6}$University of Liverpool, 
$^{7}$MD Anderson Cancer Center }\\
}

\begin{document}
\maketitle
\begin{abstract}
Understanding radiologists' eye movement during Computed Tomography (CT) reading is crucial for developing effective interpretable computer-aided diagnosis systems. However, CT research in this area has been limited by the lack of publicly available eye-tracking datasets and the three-dimensional complexity of CT volumes. To address these challenges, we present the first publicly available eye gaze dataset on CT, called \dataset, captured from expert radiologists. Then, we introduce \method, a novel 3D scanpath predictor designed specifically to process CT volumes and generate radiologist-like 3D fixation sequences, overcoming the limitations of current scanpath predictors that only handle 2D inputs. Since deep learning models benefit from a pretraining step, we develop a pipeline that converts existing 2D gaze datasets into 3D gaze data to pretrain \method. Through both qualitative and quantitative evaluations on \dataset, we demonstrate the effectiveness of our approach and provide a comprehensive assessment framework for 3D scanpath prediction in medical imaging.
Code and data are available at \url{https://github.com/UARK-AICV/CTScanGaze}.
\end{abstract}
\begin{figure}[t]
    \centering
    \includegraphics[width=\linewidth]{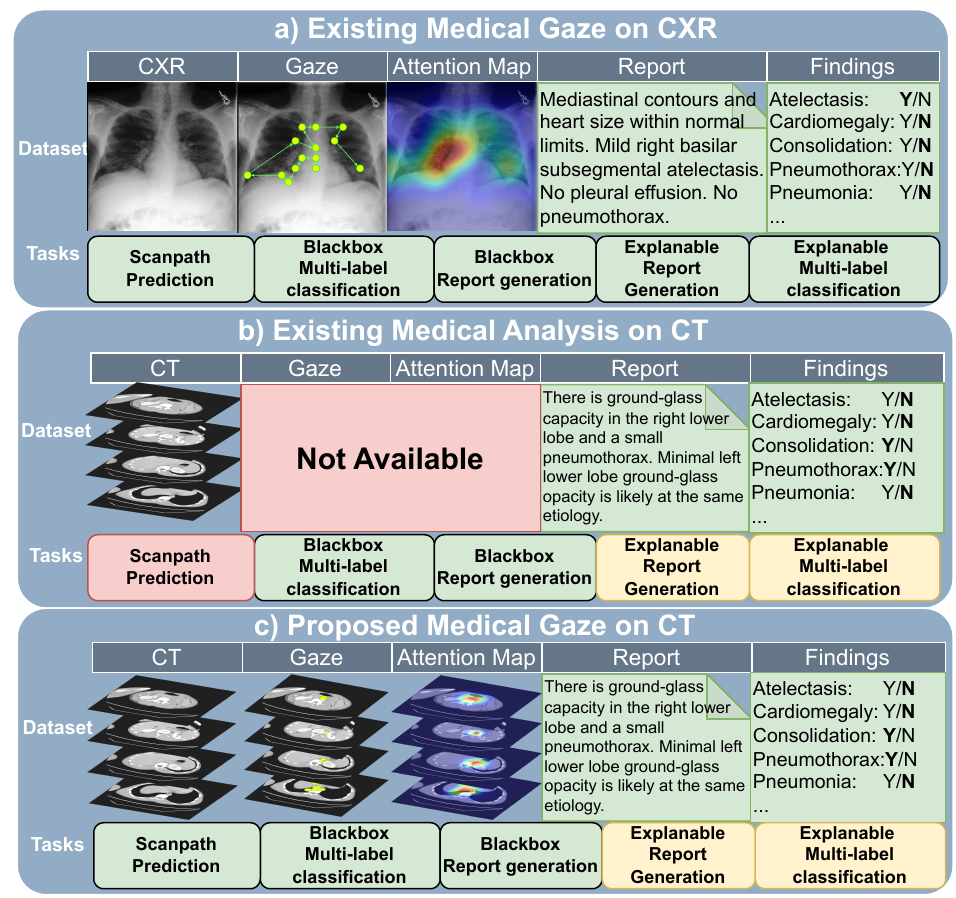}
    \caption{Many research directions in CAD would benefit from the availability of gaze, report, and findings data, as is the case for CXR (a), highlighted in green. However, some critical areas in CT research are underexplored. For example, there are only preliminary results for Explainable Report Generation and Explainable Classification, highlighted in yellow. Especially, Scanpath Prediction is underexplored and often overlooked, highlighted in red, primarily due to the lack of publicly available datasets (b). Our dataset offers new research opportunities to these tasks. In this paper, we address the Scanpath Prediction task on CT scans (c).}
    \label{fig:overview}
\end{figure}
\begin{figure*}[t]
    \centering
    \includegraphics[width=0.91\textwidth]{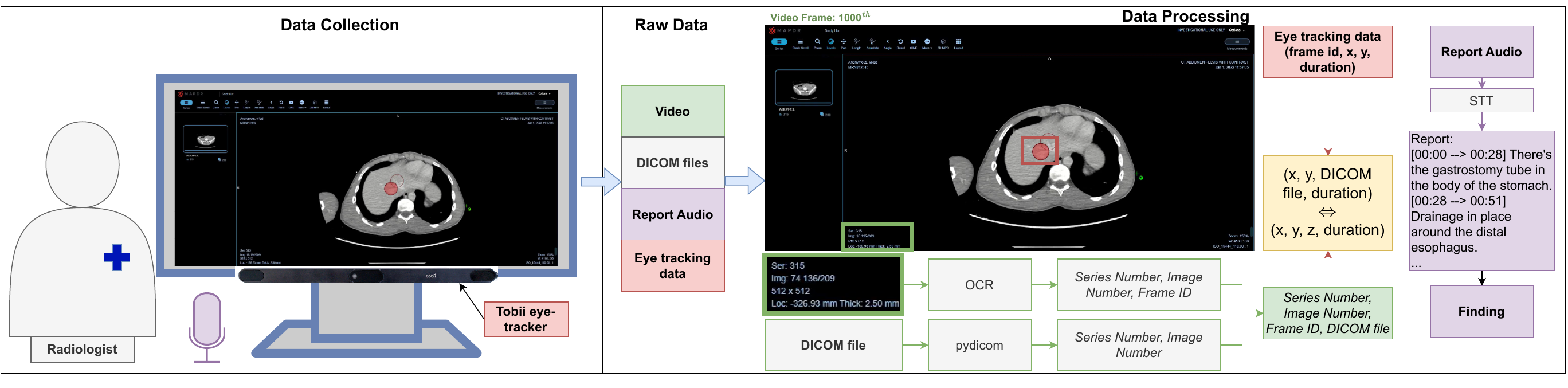}
    \caption{Illustration of our data collection and processing pipeline. The Data Collection panel shows the setup: a radiologist examines CT scans on a monitor equipped with a Tobii eye-tracker and microphone for recording audio reports. The Raw Data panel displays four data streams collected: screen recording video, DICOM files, verbal report audio, and eye gaze data.  The Data Processing panel demonstrates the data processing pipeline, which includes OCR and pydicom processing of DICOM files, integration of eye gaze data (frame ID, (x, y) coordinates, duration) to create the final 3D gaze data. We create the final radiology report and clinical findings from the Report Audio.}
    \label{fig:data_collection}
\end{figure*}
\section{Introduction}
\label{sec:intro}

Interpretability is a fundamental aspect of Computer-aided Diagnosis (CAD) system as it supports safe, accurate, and trustworthy patient care~\cite{ma2023eyegazeguidedvisiontransformer}. 
The integration of eye gaze signals offers a promising approach to enhance the interpretability of CAD systems~\cite{drew2013scannersdrillerscharacterizinga,ma2023eyegazeguidedvisiontransformer,neves2024shedding,ma2024eyegazeguidedmultimodal,karargyris2021creation}. 
Understanding the importance of gaze data in medical imaging analysis, EGD~\cite{karargyris2021creation} and REFLACX~\cite{bigolin2022reflacx} are created and shared publicly to advance chest X-rays (CXR) analysis. 
As shown in \cref{fig:overview}a, these datasets have facilitated developments for artificial intelligent solutions, including multilabel classification~\cite{karargyris2021creation,rong2021human,wang2024gazegnn}, report generation~\cite{karargyris2021creation,ramirez2022medical}, explainable classification~\cite{pham2024ai}, explainable report generation~\cite{peng2024eye}, and scanpath prediction~\cite{verma2024artificially}. 
However, existing medical datasets are either limited to 2D images like CXRs or lack radiologist gaze data. 
The absence of public dataset capturing radiologists' eye gaze patterns on Computed Tomography (CT) has left several critical tasks underexplored in CT imaging, including scanpath prediction, explainable diagnosis (classification and report generation), as shown in \cref{fig:overview}b. Especially, the scanpath prediction task on CT data is often overlooked due to missing gaze data.

The volumetric nature of CT data reveals several unique behaviors from radiologists' eye movement. For example, radiologists must constantly navigate across multiple slices and viewing planes to understand anatomical relationships~\cite{drew2013scannersdrillerscharacterizinga,alexander2022mandatinglimitsworkload}. 
Additionally, radiologists mentally integrate spatial information across different depths to form a cohesive 3D understanding of the anatomy and pathology~\cite{alexander2022mandatinglimitsworkload,eckstein2017role,drew2013informaticsradiologywhat}. 
Finally, they employ systematic search strategies to ensure thorough volume examination, as overlooking even a single slice risks missing critical findings~\cite{alexander2022mandatinglimitsworkload,drew2013scannersdrillerscharacterizinga,waite2017interpretiveerrorradiology}. 
The existing scanpath prediction methods ~\cite{yang2024hat,sounak:2023:gazeformer,chen2024isp} are primarily designed for 2D imaging analysis. 
Consequently, these 2D-based models do not account for the unique aspects of CT interpretation, particularly the complex volumetric eye movement strategies and comprehensive slice coverage patterns exhibited by radiologists, such as moving back-and-forth between slices.
Motivated by these challenges, this paper introduces the first public eye gaze medical dataset that focuses on CT scans, \dataset. Unlike existing medical eye gaze datasets~\cite{bigolin2022reflacx,karargyris2021creation},  \dataset provides 3D eye gaze data associated with every CT volume. As shown in \cref{fig:overview}c, \dataset provides four main modalities: CT scans, eye gaze data (gaze map and attention map), radiology reports, and findings. 

To conduct a benchmark on the proposed \dataset, we tackle the scanpath prediction task. 
While existing scanpath prediction methods can be extended to 3D, their original 2D design may limit their ability to model inter-slice navigation and spatial-temporal continuity. Moreover, scaling these methods to 3D introduces a higher-dimensional search space, making them more susceptible to the curse of dimensionality and harder to generalize without specialized architectural design.
Therefore, we propose a transformer-based network, \method, that generates 3D scanpaths for CT volumes. DL typically requires large pretraining datasets, but \dataset is relatively small with only 909 CT volumes compared to COCO-Search18's 6,202 images, potentially leading to overfitting or suboptimal training.
So, we utilize a 2D-to-3D pipeline to create a synthetic 3D gaze dataset for the pretraining step. By pretraining \method, it gains the ability to process CT features and predict gaze-like sequences, which enhances its performance on the final scanpath prediction task.  
Next, we train and evaluate \method against current state-of-the-art scanpath prediction models~\cite{yang2024hat,sounak:2023:gazeformer,chen2024isp}, which are adapted to process 3D inputs and predict 3D scanpaths. Our findings indicate that \method successfully generates radiologist-like scanpaths, effectively capturing both spatial coverage within each CT slice and navigational movements across slices.


Our contributions are summarized as follows:
\begin{itemize}
    \item \textbf{\dataset:} We present the first public dataset of expert radiologist gaze during CT analysis, comprising CT scans, eye gaze data, detailed reports, and findings.
    \item \textbf{\method:} We introduce a 3D scanpath prediction network that generates radiologist-like eye movement from a CT volume. And a pretraining pipeline on synthetic 3D gaze data (CT, 3D gaze) from 2D gaze data (CXR, 2D gaze).
\end{itemize}

\section{\dataset}
\label{sec:dataset}
\begin{figure*}[t]
    \centering
    \includegraphics[width=.95\textwidth]{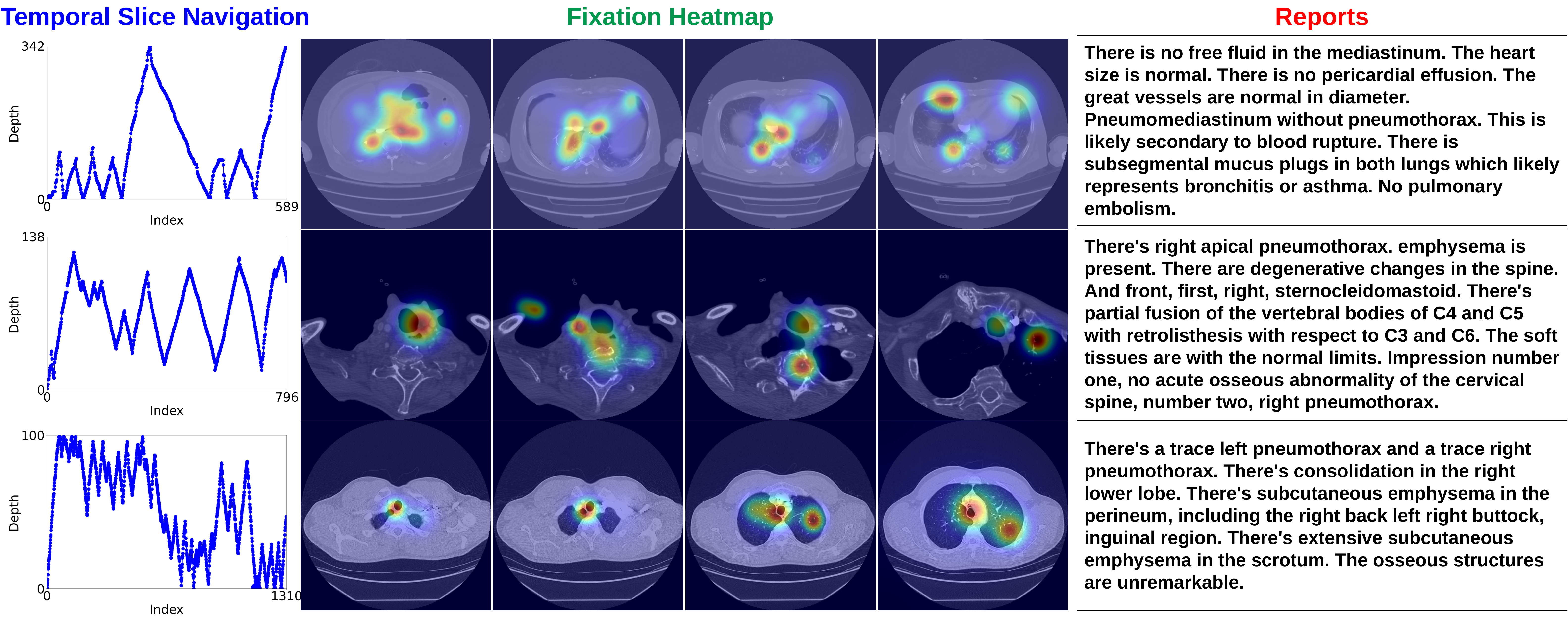}
    \caption{Examples from our dataset. Three CTs are reviewed and concluded with radiology reports as shown in the {\color{red} \textbf{Reports}} column on the right. The {\color{blue} \textbf{Temporal Slice Navigation}} shows how slice navigation change over time. We can observe that radiologists often scan through all slices and go back-and-forth for suspicious areas. In the middle column {\color{green_2} \textbf{Fixation Heatmap}}, we show some slices in the CTs. The heatmap represents all fixations viewed on that slice. For example, if in the first view they only see the left side, and in the second view they see the right side, the visualized heatmap shows fixations on both sides.}
    \label{fig:dataset-showcase}
\end{figure*}

\subsection{Data Collection}
Our study collects data from a private hospital dataset of chest and abdomen CT scans, working with two experienced radiologists (10+ years experience). 
Data collection setup involves a Tobii eye tracker mounted on a monitor to track gaze data and a microphone to record audio reports. Calibration is performed before each session, with Tobii Pro Lab software managing the eye-tracking data and screen recordings. CT scans are viewed using OHIF viewer integrated with OHIF-Orthanc server, providing necessary tools, e.g. contrast window control, for radiological analysis. We organize the setup as in \cref{fig:data_collection} (left).
Our radiologists perform their natural reading process of every CT scan exactly like they do in clinical practice. For example, 5mm images are sufficient for many pathologies. It is impractical for radiologists to review too many images in the thin series. Thin series are used for troubleshooting when radiologists need to examine certain details. Therefore, our CT scans vary in slice thickness (1-5 mm) to match this practice.



\subsection{Data Processing}
Each data collection session yields four data streams: video recording of radiologists' CT reading session, DICOM files of the CT scan, audio recording report, and eye gaze data. As shown in \cref{fig:data_collection} (right), we process these streams to create through three main steps: Spatial Mapping, Spatiotemporal Mapping, Radiological Report \& Findings Extraction. \cref{fig:dataset-showcase} shows 3 examples being viewed by our radiologists.

\noindent
\textbf{Spatial Mapping.} We extract individual frames from session videos and use Tesseract OCR~\cite{smith2007ocr} to identify series and slice numbers from the bottom-left text, with manual correction when needed. This creates frame-to-DICOM mapping pairs (\cref{fig:data_collection}).

\noindent
\textbf{Spatiotemporal Mapping.} The eye tracker records time (ms), screen coordinates $(x,y)$, and fixation duration $(t)$. Each timestamp is mapped to a frame ID using the video's frame rate (25 FPS). Using frame-to-DICOM pairs, we map gaze data to DICOM files to obtain (DICOM ID, $(x,y,t)$) pairs. Since each DICOM represents a CT slice number, we transform fixations into a list of 4-tuples $(x,y,z,t)$, where $z$ is a slice number.

\noindent
\textbf{Radiological Report \& Findings Extraction.}
We use Google’s Speech-to-Text `medical dictation' model~\cite{google_speech_to_text} on the recorded audio to generate a textual report of the radiologist’ verbal interpretation for each CT scan. From these reports, we use SARLE~\cite{draelos2021sarle}, a specialized labeler for extracting findings from CT reports, to extract the corresponding radiological findings.

\subsection{Dataset Statistics}
\label{sec:data-stats}
\dataset contains 909 CT scans, each accompanied by: scanpath, a radiology report, and findings. We have a total of 131,618 CT slices, 4,772 minutes of scanpath data, and 9,332 extracted findings. Due to the original gaze data containing dense and complex scanpaths with sequences averaging 543 fixations (and reaching up to 2,708 fixations per CT), we employ a simplification algorithm from the MultiMatch toolbox~\cite{dewhurst2012depends} to make the data more manageable while preserving essential gaze patterns. This process reduces sequences to an average of 222 fixations with a maximum of 1,507 fixations.
Both original and simplified versions will be made available.
We recommend the reader to see \cref{sec:add-stats} for more statistical details and \cref{sec:mm} for a discussion on the gaze simplification algorithm.


\subsection{Scientific Benefits}
This dataset represents a valuable resource for the medical imaging and computer vision communities:
\begin{itemize}
    \item \textbf{Benchmark for 3D Scanpath Prediction:} \dataset serves as a benchmark for predicting visual search patterns in 3D medical volumes, addressing difficulties unique to volumetric imaging that existing 2D-focused datasets do not cover.
    \item \textbf{Advancement of Explainable AI}: By linking radiologists' gaze with radiology diagnosis, \dataset supports research in explainable report generation and classification, enhancing insights into the connection between visual attention and diagnostic reasoning. This research area is actively explored in 2D medical imaging analysis by various researchers~\cite{karargyris2021creation,bigolin2022reflacx,pham2025itpctrl,wang2022follow,ibragimov2024use}.
\end{itemize}
In addition, we believe \dataset can be used in other scenarios such as study of radiologists' gaze behavior~\cite{drew2013scannersdrillerscharacterizinga}, advancement of general 3D scanpath prediction, radiology training~\cite{karargyris2021creation,wang2022follow,ibragimov2024use}, and collaborative CAD~\cite{neves2024shedding,pedrosa2020lndetector,khosravan2019collaborative}.  
\vspace{-0.41em}
\begin{figure*}[t]
    \centering
    \includegraphics[width=0.95\linewidth]{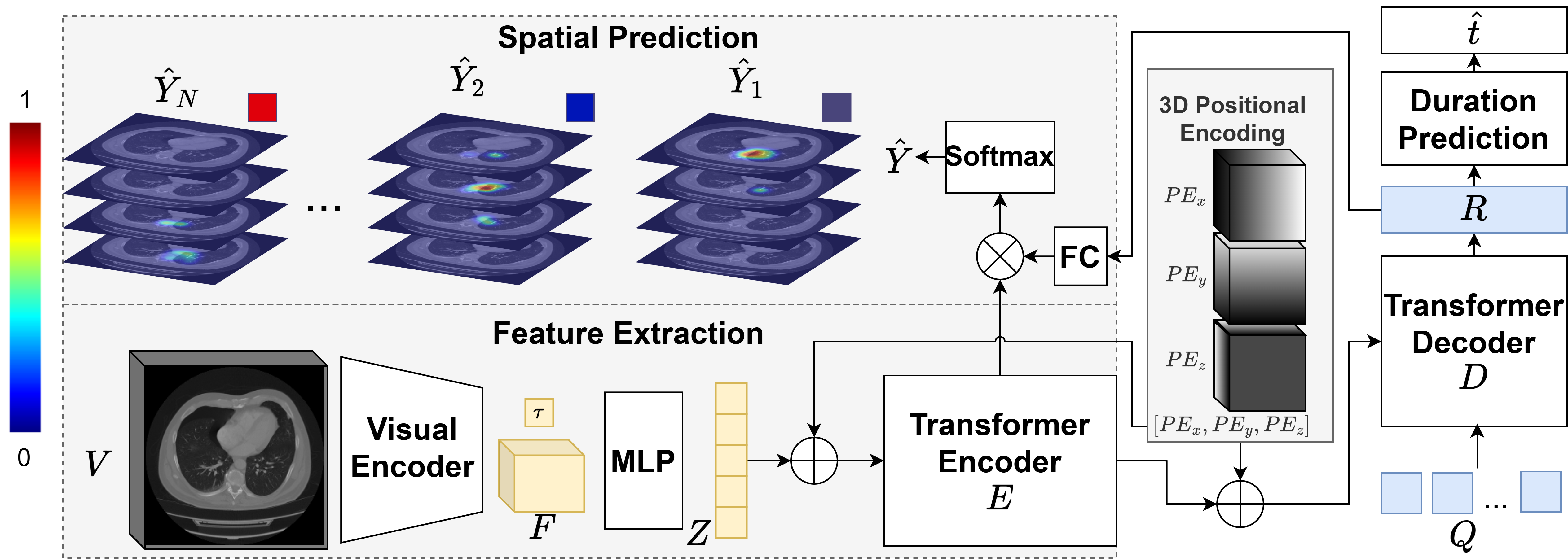}
    \caption{\method processes CT scans $V$ to predict 3D scanpaths. Initially, a 3D visual encoder within the Feature Extraction module extracts voxel features $F$. These features with a special `stop' token are then transformed into $Z$ by an MLP and combined with 3D positional encoding $PE$ to incorporate spatial information. The Transformer Encoder subsequently composes suitable 3D-aware representations from $Z$. A Transformer Decoder then uses a set of learnable queries $Q$ to attend to these 3D-aware representations and generate decoded features $R$ for the desired outputs. To prevent the loss of 3D spatial information, we reapply the 3D positional encoding before feeding the 3D-aware representations into the Transformer Decoder. The decoded features $R$ are then processed by the Spatial Prediction to produce $\hat{Y}$, including 3D fixation maps and a `stop' probability. $R$ is also processed by the Duration Prediction Head to estimate fixation durations $\hat{t}$ over time. Each $\hat{Y}$ has $H * W * D + 1$ elements, where $H * W * D$ tokens represent the 3D fixation map and one special element, embedded from token $\tau$, represents `stop'. In the final $\hat{Y}_N$ of this figure, the special token has the highest probability and is likely to be chosen in sampling, signaling that the model predicts fixation stops at the $N^{\text{th}}$ fixation.}
    \label{fig:architecture}
\end{figure*}

\section{Problem Statement: 3D Scanpath Prediction}
\label{sec:problem}
In this work, we address the 3D scanpath prediction task on CT scans.
Given a CT volume $V \in \mathbb{R}^{H \times W \times D}$, our goal is to predict a sequence of $N$ fixations $\{\hat{p}_1, \hat{p}_2, \dots, \hat{p}_N\}$. 
Each predicted fixation $\hat{p}_i = (\hat{x}_i, \hat{y}_i, \hat{z}_i, \hat{t}_i)$ should minimize its deviation from the ground truth fixation sequence $\{p_1, p_2, \dots, p_N\}$, where each $p_i = (x_i, y_i, z_i, t_i)$ represents the 3D spatial location $(x_i, y_i, z_i)$ and duration $t_i$ of a radiologist's gaze point. Our model aims to capture both the spatial patterns and temporal dynamics of expert visual search behavior in volumetric medical imaging.

\section{\method}
\label{sec:arch}
The overall architecture of \method is depicted in \cref{fig:architecture}. First, a Feature Extraction module takes a CT volume $V$ and produces suitable 3D-aware representations (\cref{sec:feature_extraction}). Then, a Transformer Decoder uses a set of learnable queries to attend on the 3D-aware presentation and creates appropriate decoded features for decoding our desired outputs (\cref{sec:transformer_decoder}). Finally, the latent features go through a Spatial Prediction module to produce 3D probability maps (\cref{sec:spatial_head}) and a Duration Prediction module to produce durations (\cref{sec:duration_head}). All modules in \method are trained jointly with losses defined in \cref{sec:losses}.

\subsection{Feature Extraction}
\label{sec:feature_extraction}
Given an input CT volume $V \in \mathbb{R}^{H \times W \times D}$, we first extract visual features $F \in \mathbb{R}^{H' \times W' \times D' \times C}$ using a visual encoder, where $H'$, $W'$, and $D'$ are the reduced spatial dimensions and $C$ is the feature dimension. These features are then projected into a compatible representation $Z = \text{MLP}(cc(F,\tau)) \in \mathbb{R}^{ L \times D_m}$ for the Transformer Encoder by a Multi-Layer Perceptron (MLP), where $cc(\cdot,\cdot)$ is a concatenate operation, $\tau$ is a special learnable token representing `stop' fixation, $D_m$ is the transformer hidden dimension and $L = H' * W' * D'+1$.
To incorporate spatial information, we adapt the 2D sinusoidal positional encoding~\cite{carion2020end} to a 3D version. For each dimension $(x,y,z)$ in a 3D volume, we generate position embeddings using different frequency bands:
\begin{equation}
    \omega_k = \exp\left(-k\frac{\log(T)}{d/2}\right), \quad k = 0,\ldots,d/2-1
\end{equation}
where $T=10000$ is the temperature parameter and $d = D_m / 3$ is the embedding dimension per axis. For each spatial position $pos$ along each axis, we compute:
\begin{align}  
    PE_{axis}(pos) &= \notag
    (\sin(pos \cdot \omega_0),\cos(pos \cdot \omega_0),\ldots, \notag\\ 
    &\sin(pos \cdot \omega_{N-1}),\cos(pos \cdot \omega_{N-1}))
\end{align}
where $N$ is the maximum length of fixation, $axis \in \{x,y,z\}$ and positions $pos$ are normalized to $[0,2\pi]$. The final encoding $PE(x,y,z)$ concatenates these components:
\begin{equation}
    PE(x,y,z) = [PE_x(x); PE_y(y); PE_z(z)] \in \mathbb{R}^{D_m}
\end{equation}
We apply 3D positional encoding on only the spatial tokens $H' * W' * D'$ of $Z$ and pass $Z$ to a Transformer Encoder that composes suitable 3D-aware representations $E(Z) \in \mathbb{R}^{L \times D_m}$ for the Transformer Decoder in the next step.

\subsection{Transformer Decoder}
\label{sec:transformer_decoder}
Before going through Transformer Decoder, we apply 3D positional encoding again on $E(Z)$ to retain the 3D spatial information. Then, a Transformer Decoder $D(\cdot)$ uses $N$ learnable gaze queries $Q \in \mathbb{R}^{N \times D_m}$ to attend to relevant features in $E(Z)$ to produce decoded features $R = D(Q, E(Z)) \in \mathbb{R}^{N \times D_m}$.

\subsection{Spatial Prediction}
\label{sec:spatial_head}
The objective of Spatial Prediction module (SP) is to generate 3D fixation maps that represent the probabilistic distribution of 3D fixation coordinates over time and the the likelihood of sequence termination. 
The output of the transformer's decoder $R \in \mathbb{R}^{N \times D_m}$ is projected into a fixation embedding by a Fully-connected (FC) layer, which is then convolved with the encoded feature map $E(Z)$ and a softmax layer to get the 3D spatial distribution for all fixations $\hat{Y} \in [0,1]^{N \times L}$:
\begin{equation}
    \hat{Y} = \text{softmax}( \text{FC}(R) \otimes E(Z)^{\top}),
    \label{eq:spatial-prediction}
\end{equation}
where $\otimes$ denotes the matrix multiplication.

\subsection{Duration Prediction}
\label{sec:duration_head}
The objective of Duration Prediction module (DP) is to generate fixation durations that reflect the probabilistic distribution of fixation durations over time.
Here, we use the re-parameterization trick~\cite{doersch2016tutorial} on the encoded features $R$ to regress them into mean values $\mu_{t}$ and log-variances $\lambda_{t}$:
\begin{align}
    \mu_{t} & = \text{MLP}_{\mu_t}(R), \quad \lambda_{t} = \text{MLP}_{\lambda_t}(R) \\
    \hat{t} & = \mu_{t} + \epsilon_{t} \cdot \exp(0.5 \lambda_{t}),
\end{align}
where $\epsilon_{t} \sim \mathcal{N}(0, 1)$ is a noise term that give our predictions a probabilistic characteristic.

\subsection{Losses}
\label{sec:losses}
To model the probabilistic nature of human fixations, we use two complementary loss functions. For the SP, which generates 3D fixation probability maps, we use Cross Entropy loss between the predicted distribution $\hat{Y}_i$ and ground truth fixation map $Y_i \in [0,1]^{L_{gt}}$ at each step $i$:
\begin{equation}
    \mathcal{L}_{ce} = \frac{-1}{N} \sum_{i=1}^{N} \sum_{l=1}^{L_{gt}} Y_i(l) \log \hat{Y}_i(l)
\end{equation}
where $L_{gt} = H*W* D + 1$ are the resolution of the ground truth volumes. $\hat{Y}_i$ is interpolated to same shape $Y_i \in \mathbb{R}^{ H * W * D + 1}$ before computing loss. The i-th ground truth 3D map $U_i$ is initialized as zero, with the fixation location $(x,y,z)$ as 1:
\begin{equation}
U_i(x',y',z') = \begin{cases}
        1 & \text{if } x' = x, y' = y, z' = z \\
        0 & \text{otherwise}
    \end{cases}
\end{equation}
where $x' \leq W, y' \leq H, z' \leq D$. We use 1-hot ground-truth $Y$ framing as classification task, and CTSearch predicts distribution $\hat{Y}$ with softmax in \cref{eq:spatial-prediction}. Because $N$ is fixed, we use padding on sequences with length less than $N$ and set the stop token $\tau$ to be $1$ and $U_i = 0$, otherwise $\tau = 0$. The ground truth heatmap is $Y_i = cc(U_i,\tau)$.

The output of DP module is a scalar, making it suitable for a regression objective. We observe that the $L_1$ loss performs effectively in our scenario.
\begin{equation}
    \mathcal{L}_{t} = \frac{1}{N}\sum_{i=1}^{N_p} \|\hat{t}_i - t_i\|_1
\end{equation}
where $t_i$ is the ground truth duration at step $i$. The final loss combines both terms:
\begin{equation}
    \mathcal{L} = \mathcal{L}_{t} + \mathcal{L}_{ce}
\end{equation}

\subsection{Pre-training \method}
\label{sec:synthetic}
Due to the high complexity of 3D scanpath prediction, we find that pretraining \method before training on \dataset is necessary to enhance its ability to process CT features and predict fixation-like sequences.

First, we combine the eye gaze data from EGD~\cite{karargyris2021creation} and REFLACX~\cite{bigolin2022reflacx}, which brings more than 3,000 pairs of CXR and fixations. We also remove invalid fixation data described by the authors~\cite{karargyris2021creation,bigolin2022reflacx} of EGD and REFLACX.

Secondly, we use a CXR-to-CT method~\cite{kyung2023perx2ct} to convert all CXRs to CTs. 
To transform 2D gaze points into 3D coordinates, we process each normalized 2D fixation set $\{(x_i, y_i, t_i)\}_{i=1}^n$, where $x_i, y_i \in [0,1]$ represent normalized coordinates in the CXR image plane (with $(0,0)$ at the top-left corner and $(1,1)$ at the bottom-right corner), and $t_i$ is the fixation duration. We perform the following steps: flip the $x$ dimension using $1-x_i$ to account for the right-to-left reading pattern in CXR; set the middle slice position as $0.5$ (where $0$ represents the first slice and $1$ represents the last slice in normalized coordinates); and map $y_i$ directly to the slice dimension while preserving the normalized scale. The middle slice position ($0.5$) is chosen because radiologists exhibit a center bias when reading axial slices. 
This process transforms the 2D fixation set $\{(x_i, y_i, t_i)\}_{i=1}^n$ into a 3D fixation set $\{(1-x_i, 0.5, y_i, t_i)\}_{i=1}^n$ with approximately 3,000 samples.

\begin{figure*}[t]
    \centering
    \includegraphics[width=.98\textwidth]{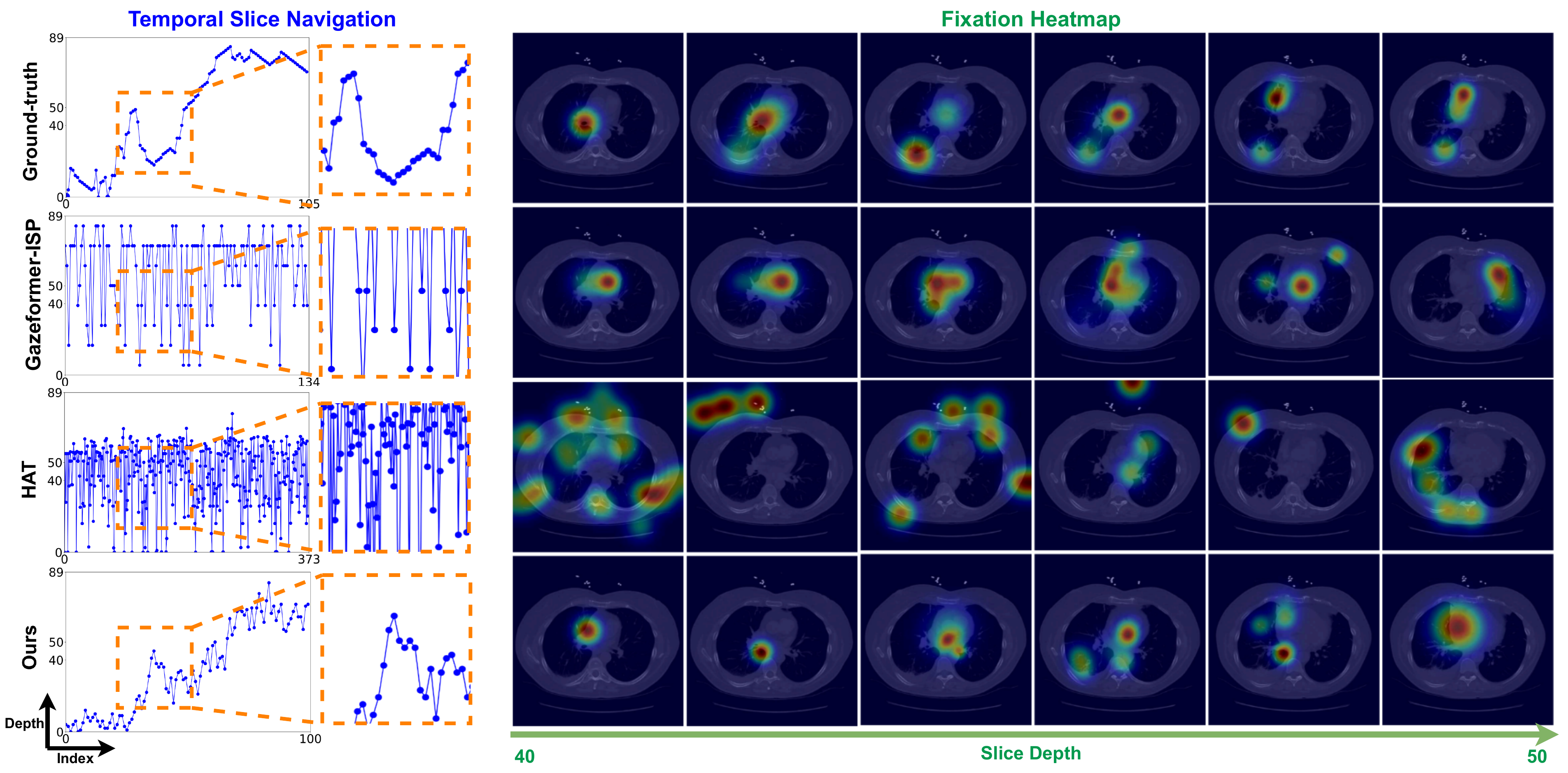}
    \vspace{-1em}
    \caption{Qualitative comparison of our method and other SOTA methods, HAT and GazeformerISP. The \textcolor{blue}{\textbf{Temporal Slice Navigation}} column shows the predicted scanpath over time, with the y-axis representing slice numbers (out of 89 total) and the x-axis showing fixation indices. For example, the ground truth has 105 fixations, with fixations move through depth from the earliest to the last slices. The \textcolor{green_2}{\textbf{Fixation Heatmap}} columns present the fixation heatmaps, illustrating regions of interest from which the corresponding gaze positions are sampled. The visualized CT slices for these heatmaps are selected between slice numbers 40 and 50. We observe that \method can capture the navigation pattern of radiologists. Especially, the \textcolor{orange}{\textbf{orange area}} indicates that the back-and-forth scanning behavior of radiologists has been successfully captured by \method.}
    \label{fig:qual}
\end{figure*}

Finally, we augment the data during pretraining by sampling fixations from a Gaussian distribution with $\sigma$ of one degree of visual angle, which fluctuates coordinates in the first and second dimension.  This follows the widely accepted assumption~\cite{lemeur2013methodscomparingscanpaths} that fixations follow a Gaussian distribution.
We emphasize that this conversion process should only be used during pretraining due to the inherent domain gap between 2D and 3D imaging. We provide additional synthetic visualization in \cref{sec:add-synthetic}. 


\begin{table*}[t]
    \centering
    \caption{Comparison of scanpath-based metrics between our \method and existing models adapted to 3D data. (\cref{sec:add-baseline}).}
    \label{tab:scanpath-comparison}
    \resizebox{\linewidth}{!}{
        \begin{tabular}{l|cc|ccccc|c}
            \toprule
            \multirow{2}{*}{\textbf{Method}} & \multicolumn{2}{c|}{\textbf{ScanMatch} $\uparrow$} & \multicolumn{5}{c}{\textbf{MultiMatch} $\uparrow$} & \multicolumn{1}{|c}{\multirow{2}{*}{\textbf{SED} $\downarrow$}} \\ \cmidrule{2-8}
             & \multicolumn{1}{c}{w/o Dur.} & \multicolumn{1}{c|}{w/ Dur.} & \multicolumn{1}{c}{Vector} & \multicolumn{1}{c}{Direction} & \multicolumn{1}{c}{Length} & \multicolumn{1}{c}{Position} & \multicolumn{1}{c|}{Duration} & \multicolumn{1}{|c}{} \\ \midrule
             PathGAN~\cite{Assens2018pathgan}& 0.0118$\pm$0.002 & 0.0649$\pm$0.005 & 0.8277$\pm$0.023 & 0.3194$\pm$0.015 & 0.6786$\pm$0.031 & 0.6559$\pm$0.028 & 0.2959$\pm$0.018 & 663$\pm$42 \\
             HAT~\cite{yang2024hat} & 0.0171$\pm$0.003 & - & 0.8103$\pm$0.019 & 0.3178$\pm$0.014 & 0.6522$\pm$0.029 & 0.6295$\pm$0.025 & - & 307$\pm$28 \\
             Gazeformer~\cite{sounak:2023:gazeformer} & 0.0619$\pm$0.005 & 0.0718$\pm$0.006 & 0.8653$\pm$0.021 & 0.3012$\pm$0.016 & 0.8601$\pm$0.035 & 0.6492$\pm$0.027 & 0.3254$\pm$0.019 & 279$\pm$25 \\
             GazeformerISP~\cite{chen2024isp} & 0.0828$\pm$0.007 & 0.0711$\pm$0.006 & 0.8831$\pm$0.024 & 0.3060$\pm$0.015 & 0.8044$\pm$0.033 & 0.7354$\pm$0.031 & 0.3375$\pm$0.020 & 238$\pm$21 \\ \midrule
             \textbf{Ours \method} & \textbf{0.1466$\pm$0.009} & \textbf{0.1170$\pm$0.008} & \textbf{0.9216$\pm$0.026} & \textbf{0.4151$\pm$0.018} & \textbf{0.8783$\pm$0.036} & \textbf{0.7859$\pm$0.033} & \textbf{0.5003$\pm$0.024} & \textbf{174$\pm$18} \\ \bottomrule
        \end{tabular}
    }
    \label{tab:main-comparison1}
\end{table*}

\begin{table}[b]
    \centering
    \vspace{-1em}
    \caption{Comparison of spatial-based metrics between different scanpath prediction models.}
    \label{tab:saliency-comparison}
    \setlength{\tabcolsep}{6.0pt}
    \resizebox{0.97\linewidth}{!}{
        \begin{tabular}{l|ccc}
            \toprule
            \multirow{2}{*}{\textbf{Method}} & \multicolumn{3}{c}{\textbf{Saliency}} \\ \cmidrule{2-4}
             & CC $\uparrow$ & KLDiv $\downarrow$ & NSS $\uparrow$ \\ \midrule
             PathGAN~\cite{Assens2018pathgan} & 0.0349$\pm$0.017 & 25.602$\pm$1.832 & 0.1343$\pm$0.021 \\
             HAT~\cite{yang2024hat} & 0.0914$\pm$0.034 & 15.493$\pm$1.245 & 1.0676$\pm$0.089 \\
             Gazeformer~\cite{sounak:2023:gazeformer} & 0.0855$\pm$0.044 & 23.332$\pm$1.756 & 0.5005$\pm$0.042 \\
             GazeformerISP~\cite{chen2024isp} & 0.1104$\pm$0.021 & 5.023$\pm$0.428 & 0.7703$\pm$0.065 \\ \midrule
             \textbf{Ours \method} & \textbf{0.1706$\pm$0.016} & \textbf{3.645$\pm$0.312} & \textbf{1.1422$\pm$0.095} \\ \bottomrule
        \end{tabular}
    }
    \label{tab:main-comparison2}
\end{table}
\begin{table*}[t]
    \centering
    \caption{Performance of our proposed \method on different radiologists on scanpath-based metrics.}
    \label{tab:scanpath-comparison}
    \resizebox{\linewidth}{!}{
        \begin{tabular}{l|cc|ccccc|c}
            \toprule
            \multirow{2}{*}{\textbf{Method}} & \multicolumn{2}{c|}{\textbf{ScanMatch} $\uparrow$} & \multicolumn{5}{c}{\textbf{MultiMatch} $\uparrow$} & \multicolumn{1}{|c}{\multirow{2}{*}{\textbf{SED} $\downarrow$}} \\ \cmidrule{2-8}
             & \multicolumn{1}{c}{w/o Dur.} & \multicolumn{1}{c|}{w/ Dur.} & \multicolumn{1}{c}{Vector} & \multicolumn{1}{c}{Direction} & \multicolumn{1}{c}{Length} & \multicolumn{1}{c}{Position} & \multicolumn{1}{c|}{Duration} & \multicolumn{1}{|c}{} \\ \midrule
             Radiologist \#1 & 0.1499$\pm$0.010 & 0.1214$\pm$0.009 & 0.9087$\pm$0.025  & 0.4023$\pm$0.019 & 0.8888$\pm$0.035 & 0.8001$\pm$0.034 & 0.5110$\pm$0.023 & 182$\pm$19 \\
             Radiologist \#2 & 0.1431$\pm$0.008 & 0.1124$\pm$0.007 &  0.9339$\pm$0.027& 0.4273$\pm$0.017  & 0.8673$\pm$0.037 & 0.7712$\pm$0.032 & 0.4891$\pm$0.025 &   166$\pm$17\\
             Both & 0.1466$\pm$0.009 & 0.1170$\pm$0.008 & 0.9216$\pm$0.026 & 0.4151$\pm$0.018 & 0.8783$\pm$0.036 & 0.7859$\pm$0.033 & 0.5003$\pm$0.024 & 174$\pm$18 \\ \bottomrule
        \end{tabular}
    }
    \label{tab:sub-rad-cross1}
\end{table*}

\begin{table}[b]
    \centering
    \caption{Performance of \method on different radiologists on spatial-based metrics.}
    \label{tab:saliency-comparison}
    \setlength{\tabcolsep}{6.0pt}
    \resizebox{\linewidth}{!}{
        \begin{tabular}{l|ccc}
            \toprule
            \multirow{2}{*}{\textbf{Method}} & \multicolumn{3}{c}{\textbf{Saliency}} \\ \cmidrule{2-4}
             & CC $\uparrow$ & KLDiv $\downarrow$ & NSS $\uparrow$ \\ \midrule
            Radiologist \#1 & 0.1901$\pm$0.017 & 3.388$\pm$0.293 & 1.209$\pm$0.101 \\
            Radiologist \#2 & 0.1503$\pm$0.014 & 3.912$\pm$0.331 & 1.072$\pm$0.089 \\ 
             Both & 0.1706$\pm$0.016 & 3.645$\pm$0.312 & 1.142$\pm$0.095 \\ \bottomrule
        \end{tabular}
    }
    \label{tab:sub-rad-cross2}
\end{table}

\section{Experiments}

\subsection{Experimental Details}
\label{sec:experimental_detail}
\noindent 
\textbf{Implementation Details.} \method uses Swin UNETR~\cite{hatamizadeh2021swin} encoder with $96\times96\times96$ input window. Due to computational constraints, we use a frozen pre-trained Swin UNETR checkpoint on LIDC-IDRI~\cite{armato2011lung}. The transformer has 6 encoder/decoder layers, 8 attention heads, and hidden size of 768. All MLPs in \method have two layers, ReLU activation, and 512 hidden dimension. The \method is implemented in PyTorch. We use AdamW~\cite{loshchilov2017adamw} optimizer with initial learning rate of $1e^{-4}$ and weight decay of $0.01$. We train and evaluate all methods on the simplified version (\cref{sec:data-stats}) of \dataset.
We pre-train \method (\cref{sec:synthetic}) for 50,000 iterations with a batch size of 8, and we then fine-tune \method on \dataset for 20 epochs with $N=400$ fixations. For more implementation details, please refer to \cref{sec:add-implementation}. 


\noindent
\textbf{Evaluation Metrics.} We benchmark on two aspects: \textbf{scanpath-based} metrics (SM~\cite{cristino2010scanmatch}, MM~\cite{dewhurst2012depends}, SED~\cite{brandt1997spontaneous,foulsham2008can}) and \textbf{spatial-based} metrics (CC, KLDiv, NSS~\cite{huang2015salicon,chen2024isp}). All metrics are adapted for 3D scanpath prediction (\cref{sec:metrics}). We run 5-fold cross-validation and report the scores with 95\% confidence intervals.


\noindent
\textbf{Baselines.} We evaluate \method against leading scanpath prediction methods including PathGAN~\cite{Assens2018pathgan} Gazeformer~\cite{sounak:2023:gazeformer}, GazeformerISP~\cite{chen2024isp} and HAT~\cite{yang2024hat}. To benchmark these 2D methods on 3D gaze data, we replace the feature encoder of PathGAN, Gazeformer, GazeformerISP, and HAT to Swin UNETR encoder. We replace 2D positional encoding in Gazeformer, GazeformerISP, and HAT to our 3D positional encoding. Because Gazeformer has two heads for predicting the 2D coordinates and one head for predicting the duration, we only add a head to predict the slice number. In HAT and GazeformerISP, we alter their spatial decoder from generating 2D fixation heatmaps to 3D fixation heatmaps. For remaining modules, we follow their original implementation, for example we use RoBERTa~\cite{liu2019roberta} for task embedding of Gazeformer and GazeformerISP. More implementation details on 3D adapted scanpath prediction baselines are in \cref{sec:add-baseline}. 
\subsection{Qualitative Results}

We present qualitative results in Fig. \ref{fig:qual} to demonstrate the effectiveness of our proposed method. The results reveal clear performance differences across evaluation aspects. In the temporal slice navigation plots (Fig. \ref{fig:qual}, left), \method captures similar temporal dynamics to the ground-truth, while baseline methods (GazeformerISP and HAT) exhibit erratic, high-variance patterns. Notably, the \textcolor{orange}{\textbf{orange areas}} show that \method replicates the non-linear behavior characteristic of experienced readers, who frequently navigate back and forth through the volume to revisit suspicious regions. The fixation heatmap visualization (Fig. \ref{fig:qual}, right) demonstrates that \method produces more accurate attention distributions compared to other methods. These qualitative findings confirm that our approach successfully replicates the complex visual search patterns of radiologists during CT interpretation, representing a significant advancement in 3D medical scanpath prediction.
We provide additional qualitative results in \cref{sec:add-qual}.

\subsection{Quantitative Results}

\cref{tab:main-comparison1,tab:main-comparison2} demonstrates the significant challenges in 3D scanpath prediction. \method achieves consistently better performance across all metrics compared to existing approaches. The earliest method, PathGAN, performs poorly across all metrics, with particularly low ScanMatch (0.0118) and saliency scores (CC: 0.0349). Recent transformer-based approaches show improvement but face various limitations: HAT struggles with limited training data (ScanMatch: 0.0171), Gazeformer's separate coordinate prediction via three MLPs causes consistency issues affecting saliency metrics (KLDiv: 23.332), and GazeformerISP underperforms on \dataset due to its specialized and complex architecture. \method demonstrates superior performance with significant improvements in ScanMatch (0.1466), MultiMatch position (0.7859) and duration (0.5003) scores, and saliency metrics (CC: 0.1706, KLDiv: 3.645), indicating better capture of radiologists' gaze patterns. These results highlight the effectiveness of our approach and represent a substantial advancement in modeling expert visual search behavior for CT.

\begin{table}[b]
\centering
\caption{Ablation study on the impact of each component. \textbf{mSM} and \textbf{mMM} denote for ScanMatch and Multimatch, respectively.}
\label{tab:ablation}
\setlength{\tabcolsep}{9.0pt}
\resizebox{\linewidth}{!}{
\begin{tabular}{cc|cccc}
\toprule
\textbf{Pre} & \textbf{3D-PE} & \textbf{mSM} $\uparrow$ & \textbf{mMM} $\uparrow$ & \textbf{SED} $\downarrow$ & \textbf{KLDiv}$\downarrow$ \\ \hline
\xmark & \xmark & 0.0207$\pm$0.011 & 0.6034$\pm$0.031 & 252$\pm$23 & 20.958$\pm$0.902 \\
\xmark & \cmark & 0.1275$\pm$0.005 & 0.6883$\pm$0.022 & 184$\pm$18 & 5.194$\pm$0.523 \\
\cmark & \xmark & 0.0632$\pm$0.019 & 0.6562$\pm$0.019 & 189$\pm$12 & 7.526$\pm$0.616 \\
\cmark & \cmark & \textbf{0.1318$\pm$0.009} & \textbf{0.7002$\pm$0.027} & \textbf{174$\pm$18} & \textbf{3.645$\pm$0.312} \\ \bottomrule
\end{tabular}
}
\end{table}

\subsection{Cross-radiologist Evaluation}
\label{sec:cross-rad}
To assess potential learning bias toward a single radiologist's style (as \dataset is collected from only two radiologists), we conduct a cross-radiologist evaluation. Using the same trained checkpoints from our 5-fold cross validation that produce the results in \cref{tab:main-comparison1,tab:main-comparison2}, we compute separate scores for test sets containing only the first or second radiologist's ground truth data, as shown in \cref{tab:sub-rad-cross1,tab:sub-rad-cross2}. The differences in scores between the two radiologists are insignificant, leading us to conclude that \method successfully learns general scanpath patterns rather than overfitting to an individual radiologist's style.

\subsection{Ablation Studies}
We conduct comprehensive ablation studies to validate each component of our framework. We provide additional ablation study on CT Visual Encoder backbone in \cref{sec:compare-ct-backbone}. 

\noindent
\textbf{Impact of 3D Positional Encoding (3D-PE).} 3D-PE enhances the model's ability to capture positional relationships across height, width, and depth. As shown in \cref{tab:ablation}, applying 3D-PE improves our model performance across all metrics. Comparing rows \#1 and \#2 reveals substantial improvements when adding 3D-PE without pretraining, while comparing rows \#3 and \#4 demonstrates that 3D-PE remains crucial even with pretraining in place. The dramatic improvement in KLDiv from 7.526 to 3.645 between rows \#3 and \#4 highlights how 3D-PE enhances spatial awareness in our volumetric predictions.
In conclusion, 3D-PE provides essential spatial context for volumetric data processing, resulting in substantial performance improvements across all metrics regardless of whether pretraining is used.

\noindent
\textbf{Impact of Pre-training Step (Pre).} Pretraining often benefits deep learning models by establishing helpful inductive biases ~\cite{xu2021inductivebias}. Given the limited size of our real CT dataset, direct training alone may lead to suboptimal performance. As shown in \cref{tab:ablation}, our experiments demonstrate that incorporating a pretraining step further improves model performance. When comparing rows \#2 and \#4, we observe that adding pretraining to a model with 3D-PE further enhances performance, with improvements in mSM (0.1275 to 0.1318), mMM (0.6883 to 0.7002), SED (184 to 174), and KLDiv (5.194 to 3.645). These gains appear across all evaluation metrics, confirming the value of pretraining.


\section{Related Work}

\noindent 
\textbf{Eye Gaze Datasets.} 
The proliferation of eye gaze datasets in the general visual domain~\cite{jiang2015salicon,papadopoulos2014training,ehinger2009modelling,zelinsky2019benchmarking,gilani2015pet,shi:2020:air,shuo:2015:austim,juan:2014:osie,zhibo:2020:cocosearch,huiyu:2019:saliency4asd} reflects growing interest in understanding human visual behavior. These datasets span diverse scenarios, ranging from multi-target search tasks~\cite{gilani2015pet} to focused single-category search~\cite{ehinger2009modelling,zelinsky2019benchmarking}. Notable examples include COCO-Search18~\cite{zhibo:2020:cocosearch} with its extensive object categories and datasets incorporating Visual Question Answering paradigms~\cite{shi:2020:air}. While the general domain has seen substantial progress, medical eye gaze datasets remain limited in scope. Current medical datasets concentrate primarily on 2D modalities, particularly chest X-rays, as exemplified by EGD~\cite{karargyris2021creation} and REFLACX~\cite{bigolin2022reflacx}. The absence of 3D medical eye gaze datasets represents a significant gap in the field.
To address this limitation, we present the first comprehensive eye gaze dataset for CT scan analysis. \dataset provides essential data for advancing research in volumetric medical image analysis and understanding expert visual search patterns in 3D medical contexts.

\noindent 
\textbf{Scanpath Prediction.}
Early approaches to scanpath prediction primarily focused on sampling fixations from saliency maps~\cite{wei:2011:stde,olivier:2015:saccadicmodel,calden:2018:star-fc,laurent:1998:visualattention}. The field has since witnessed remarkable progress~\cite{zhang2018finding,adeli2018deep,Wei-etal-NIPS16,matthias:2016:deepgaze,marcella:2018:sam,xun:2015:salicon,camilo:2020:umsi,souradeep:2022:agdf,sen:2020:eml,shi:2023:personalsaliency,bahar:2023:tempsal,xianyu:2021:vqa,sounak:2023:gazeformer,wanjie:2019:iorroi,zhibo:2020:cocosearch,zhibo:2023:humanattention,zhibo:2022:targetabsent,ryan:2022:scanpathnet,mengyu:2023:scanpath,xiangjie:2023:scandmm,chen2024isp,yang2024hat,Pham_2024_fgcxr,gazesearch}, particularly through the integration of deep neural networks~\cite{xianyu:2021:vqa,sounak:2023:gazeformer,wanjie:2019:iorroi,zhibo:2020:cocosearch,zhibo:2023:humanattention,zhibo:2022:targetabsent,ryan:2022:scanpathnet,mengyu:2023:scanpath,matthias:2022:deepgaze,yue:2023:ueyes}, reinforcement learning~\cite{xianyu:2021:vqa,zhibo:2020:cocosearch,zhibo:2022:targetabsent}, and transformer architectures~\cite{mengyu:2023:scanpath,sounak:2023:gazeformer,yang2024hat,chen2024isp}. These advances have substantially enhanced our understanding of temporal attention dynamics. However, none of previous approaches have been designed for Computed Tomography. To address this limitation, we propose a transformer-based method, \method, specifically designed for scanpath prediction on CTs.

\section{Conclusion}
This paper introduces \dataset, the first public CT eye gaze dataset, along with \method as the first CT scanpath prediction baseline. Experiments show that \method works well and marks an important breakthrough in how we can model the way experts visually search through complex 3D medical images.

\noindent
\textbf{Acknowledgments.} This material is based upon work supported by the National Science Foundation (NSF) under Award No OIA-1946391, NSF 2223793 EFRI BRAID, National Institutes of Health (NIH) 1R01CA277739-01.

\clearpage
{
    \small
    \bibliographystyle{ieeenat_fullname}
    \bibliography{main}
}
\clearpage
\appendix
\renewcommand{\thetable}{\Roman{table}}
\renewcommand{\thefigure}{\Roman{figure}}
\setcounter{table}{0}
\setcounter{figure}{0}
\maketitlesupplementary

\section*{Ethical Statement}
This research follows all relevant ethical guidelines for medical research. All patient data used in this study was properly anonymized and de-identified following HIPAA guidelines. The radiologists who participated in the eye-tracking study provided informed consent. No personal or identifying information is included in the dataset or results. Our study aims to augment, not replace, clinical expertise and maintains the central role of human medical professionals in diagnostic decisions.

\section*{Summary}
The appendix is organized as follows:

\begin{itemize}
    \item \cref{sec:add-stats} describes additional dataset statistics including gaze data split, number of slices distribution, and duration of data collection recording videos.
    \item \cref{sec:add-synthetic} presents additional visualizations of our synthetic training data.
    \item \cref{sec:add-implementation} provides additional implementation details.
    \item \cref{sec:metrics} describes the 3D scanpath similarity metrics.
    \item \cref{sec:add-baseline} describes implementation details of baseline methods adapted for 3D scanpath prediction.
    \item \cref{sec:add-qual} provides additional qualitative results and visualizations.
    \item \cref{sec:compare-ct-backbone} compares different CT Visual Encoder backbones.
    \item \cref{sec:mm} discusses the MultiMatch simplification algorithm and analysis.
    \item \cref{sec:discussion} discuss the broader impact of our works. 
\end{itemize}

\section{Additional Dataset Details}
\label{sec:add-stats}
\subsection{Gaze Data Splits}
Our dataset consists of 909 CT-gaze pairs, split into training, validation, and test sets with a ratio of 70:10:20 respectively. This translates to:
\begin{itemize}
    \item Training set: 636 pairs
    \item Validation set: 90 pairs
    \item Test set: 183 pairs
\end{itemize}
\begin{figure*}[t]
    \centering
    \includegraphics[width=\linewidth]{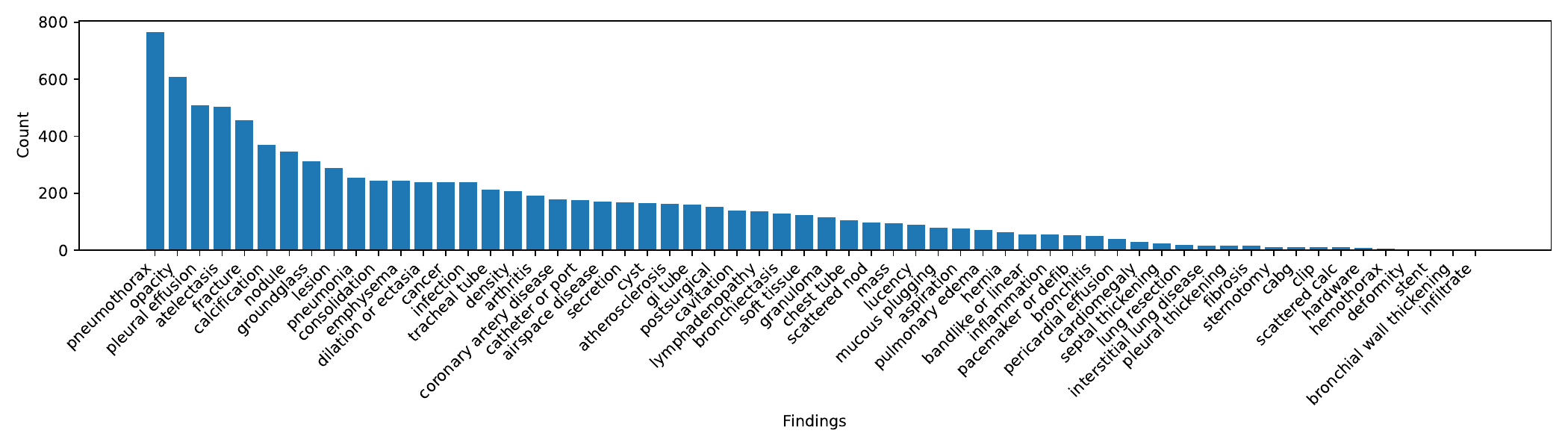}
    \caption{Extracted radiological finding histogram. The y-axis represents the findings. The x-axis represents the occurrence frequency (number of samples). From SARLE~\cite{draelos2021sarle}, we extract a total of 9,332 findings with 60 unique finding names.}
    \label{fig:rad_finding_dist}
\end{figure*}

\subsection{Radiological Finding Distribution}
\dataset has a total of 9,332 findings with 60 unique finding names. The distribution of our dataset is shown in \cref{fig:rad_finding_dist}.

\subsection{Number of CT Slices}
\begin{figure}[t]
    \centering
    \includegraphics[width=\linewidth]{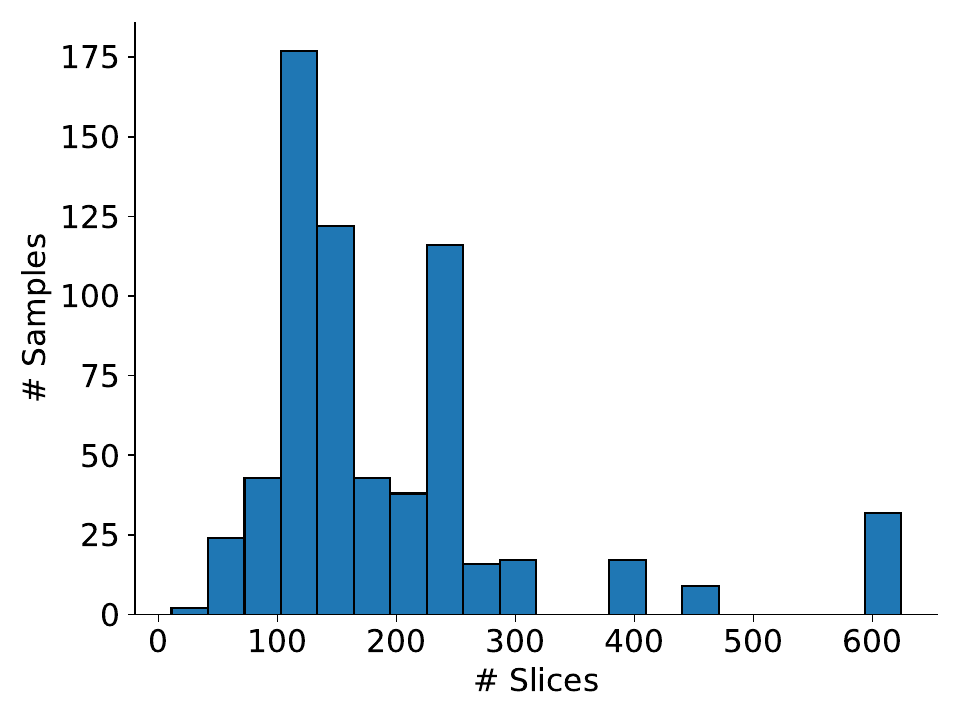}
    \caption{Number of slices histogram. The y-axis represents the number of slices. The x-axis represents the occurrence frequency (number of samples). \dataset has 131,618 slices in total and 186 slices per CT volume in average.}
    \label{fig:ct_stats}
\end{figure}
\cref{fig:ct_stats} shows the distribution of slice counts for all CT volumes. While all CT slices have a fixed resolution of $512 \times 512$ pixels in the axial plane, the number of slices varies across volumes with a total number of slices is 131,618 and 186 slices per CT in average.

\subsection{Recording Duration}
In our dataset, we prioritize maintaining natural workflow by allowing radiologists to read CT scans following their standard clinical practice. Table~\ref{tab:recording_stats} summarizes the duration statistics of our video recordings. 
\begin{table}[h]
    \centering
    \caption{Recording Duration Statistics.}
    \begin{tabular}{lc}
        \hline
                             & Duration (minutes) \\
        \hline
        Total recording time & 4722                \\
        Average session time & 5.36                  \\
        Minimum session time & 1.27               \\
        Maximum session time & 9.68                 \\
        \hline
    \end{tabular}
    \label{tab:recording_stats}
\end{table}
\subsection{Additional Visualization of \dataset}

\begin{figure*}[t]
    \centering
    \includegraphics[width=0.9\linewidth]{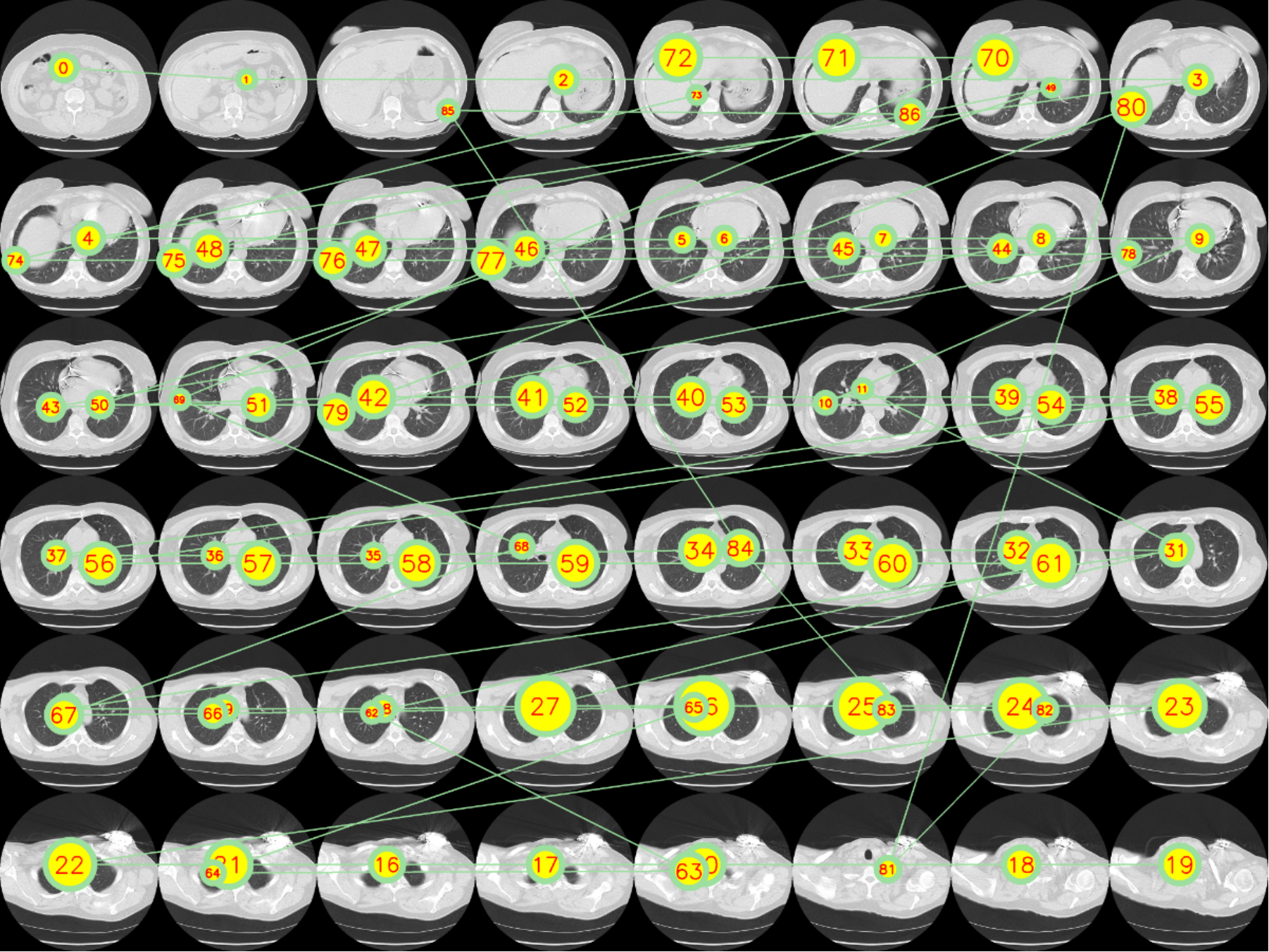}
    \caption{Illustration of fixation sequence across sequential CT slices, following a left-to-right and top-to-bottom order. The fixation start from $0$ in the top left corner. The numbered annotations indicate the order of fixation points. In this figure, we observe the radiologist's systematic viewing pattern: first scanning through all slices before returning to central regions for detailed examination. This navigation pattern is also demonstrated in \cref{fig:dataset-grid-line}. We suggest the readers to watch \texttt{vis\_gt.mp4} to see the animated version of this figure. }
    \label{fig:dataset-grid-grid}
\end{figure*}
\begin{figure}[t]
    \centering
    \includegraphics[width=0.9\linewidth]{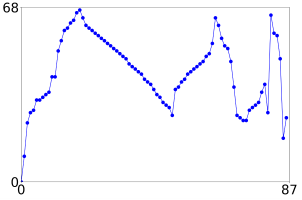}
    \caption{Visualization of temporal navigation patterns from the gaze data in \cref{fig:dataset-grid-grid}. The scanpath reveals that the radiologist follows a systematic approach: first traversing from start to end slices, then returning to central regions for detailed examination.}
    \label{fig:dataset-grid-line}
\end{figure}

We visualize a CT volume with its eye tracking data from an alternative point of view, showing all scanpaths across slices in \cref{fig:dataset-grid-grid} and temporal navigation patterns in a line chart in \cref{fig:dataset-grid-line}. To create \cref{fig:dataset-grid-grid,fig:dataset-grid-line}, we select a CT volume such that its fixation sequence has only 48 unique slices (48 unique z values) to maintain the simplicity of the visualization. An animated video is provided, named \texttt{vis\_gt.mp4}. 
The observed scanpath pattern demonstrates a natural progression from peripheral regions inward to areas of diagnostic significance. The timestamped report is:
\begin{lstlisting}
[00:00.000 --> 00:29.960]  there's a left upper chest pacemaker or ICD
[00:29.960 --> 00:40.240]  with leads in the right atrium right ventricle and a pericardial lead along
[00:40.240 --> 00:43.240]  the left ventricle
[00:43.240 --> 00:57.960]  there are no enlarged axillary or supraclavicular lymph nodes
[00:57.960 --> 01:19.320]  mildly enlarged paratracheal lymph nodes are present there's a mildly enlarged
[01:19.320 --> 01:28.320]  large lymph node in the anterior mediastinum the left ventricle appears
[01:28.320 --> 01:34.680]  mildly dilated with fatty metaplasia in the left ventricular apex and
[01:34.680 --> 01:43.320]  interventricular septum there's no pericardial effusion the great vessels
[01:43.320 --> 01:49.920]  are normal in diameter there's mild aortic atherosclerotic calcification
[01:49.920 --> 01:55.920]  there is a stent in the LAD
[02:01.720 --> 02:04.720]  there's no pleural effusion
[02:04.720 --> 02:19.840]  there is cholelithiasis a low density nodule is present in the left adrenal
[02:19.840 --> 02:25.800]  gland likely representing an adenoma there's a calcified granuloma in the
[02:25.800 --> 02:28.800]  spleen
[02:34.720 --> 02:37.720]  there's no pericardial effusion
[02:37.720 --> 02:40.720]  there's no pericardial effusion
[02:40.720 --> 03:06.200]  a small right pneumothorax is present
[03:10.720 --> 03:17.720]  the trachea and central airways are clear
[03:17.720 --> 03:31.720]  a calcified granuloma is present in the right upper lobe
[03:31.720 --> 03:38.720]  a small area of ground glass opacity is present in the right lower lobe
[03:38.720 --> 03:50.720]  there's a right lower lobe nodule measuring approximately 15 millimeters
[04:08.720 --> 04:24.440]  impression number one there's a small right pneumothorax number two a small
[04:24.440 --> 04:27.840]  area of ground glass opacity in the peripheral right lower lobe is likely
[04:27.840 --> 04:37.440]  infectious inflammatory there may be a cavitary component which could be the
[04:37.440 --> 04:43.240]  cause of the right pneumothorax number three there's a solid right lower lobe
[04:43.240 --> 04:47.960]  nodule measuring 15 millimeters the differential includes infection slash
[04:47.960 --> 04:50.960]  inflammation and malignancy
[04:54.360 --> 04:57.360]  number four
[04:57.360 --> 05:00.360]  mildly enlarged lymph nodes in the mediastinum are nonspecific
\end{lstlisting}

By removing the timestamps (e.g., \verb|[00:00.000 --> 00:29.960]|), we obtain a free-text radiology report:
\textit{
    There's a left upper chest pacemaker or ICD with leads in the right atrium right ventricle and a pericardial lead along the left ventricle. There are no enlarged axillary or supraclavicular lymph nodes. Mildly enlarged paratracheal lymph nodes are present. There's a mildly enlarged large lymph node in the anterior mediastinum. The left ventricle appears mildly dilated, with fatty metaplasia in the left ventricular apex and interventricular septum. There's no pericardial effusion the great vessels are normal in diameter there's mild aortic atherosclerotic calcification. There is a stent in the LAD there's no pleural effusion, there is cholelithiasis. A low density nodule is present in the left adrenal gland, likely representing an adenoma. There's a calcified granuloma in the spleen. There's no pericardial effusion. There's no pericardial effusion. A small right pneumothorax is present. The trachea and central airways are clear. A calcified granuloma is present in the right upper lobe. A small area of ground glass opacity is present in the right lower lobe. There's a right lower lobe nodule measuring approximately 15 millimeters. IMPRESSIONS. Number one. There's a small right pneumothorax. Number two. A small area of ground glass opacity in the peripheral right lower lobe is likely infectious inflammatory. There may be a cavitary component, which could be the cause of the right pneumothorax. Number three. There's a solid right lower lobe nodule measuring 15 millimeters, the differential includes infection slash inflammation and malignancy. Number four. Mildly enlarged lymph nodes in the mediastinum are nonspecific.
}

Using CheXbert~\cite{smit2020chexbert} to extract the 13 CheXpert findings~\cite{irvin2019chexpert}, we identify the following positive findings: `Enlarged Cardiomediastinum', `Lung Lesion', `Pleural Effusion', and `Support Devices'.

\subsection{Example of \dataset}
We also provide one example of our data in \verb|example_data.zip|.
\begin{itemize}
    \item \verb|ct_id9.nii.gz| is the CT scan. 
    \item \verb|finding_id9.csv| is the finding annotations.
    \item \verb|fixation_id9.json| is the original fixations (without being simplified).
    \item \verb|recorded_video_id9.mp4| is the recorded video session.
    \item \verb|report_id9.txt| is the report created by speech to text software. 
\end{itemize}
\section{Additional Visualization of Synthetic Data}
\label{sec:add-synthetic}
\begin{figure*}[t]
    \centering
    \includegraphics[width=\linewidth]{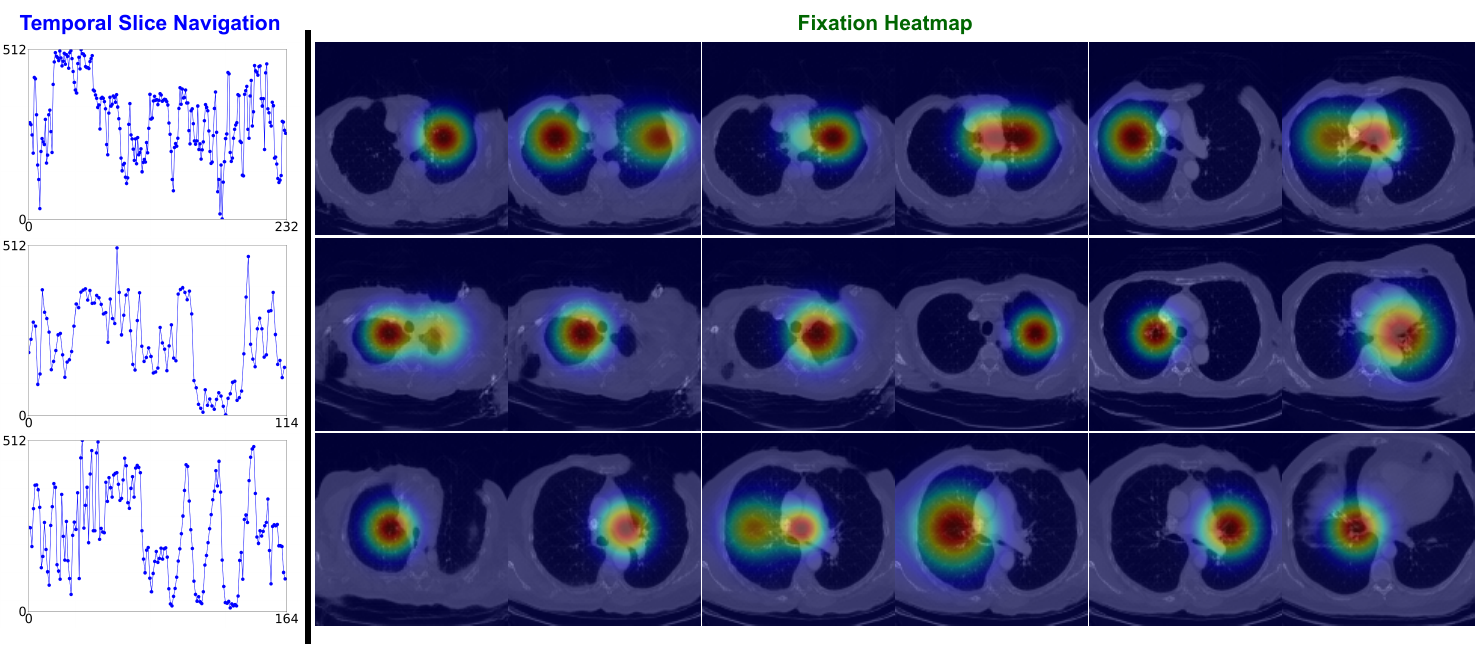}
    \caption{Visualization of our synthetic training data. The Temporal Slice Navigation demonstrates the temporal transition of slice. The Fixation Heatmap column displays representative CT slices with corresponding fixation heatmaps.}
    \label{fig:synthetic_viz}
\end{figure*}

\cref{fig:synthetic_viz} illustrates samples from our synthetic dataset, comprising temporal slice navigation patterns and fixation heatmaps. Overall, the synthetic data exhibits similarities with real eye movement for 3D, particularly in temporal characteristics when transitioning slices along the depth dimension.

\section{Additional Implementation Details}
\label{sec:add-implementation}
Due to GPU memory constraint, directly using a full CT volume as input to extract complete feature maps and train end-to-end is often not feasible. In our implementation, we follow the common practice of sliding windows and merging windows. \method uses Swin UNETR~\cite{hatamizadeh2021swin} encoder with $96\times96\times96$ input window. The output shape is reduced to $3\times3\times3$ with feature dimension $C=768$. With a downsampling ratio $r=32$, we merge all features into a single feature map with shape $(W', H', D') = (W/r, H/r, D/r)$. For technical convenience, we interpolate all feature maps to a standardized size of $16\times16\times16$ with $C=768$ channels, ensuring uniformity across varying feature map shapes. This standardization is reasonable since our CT scans have fixed dimensions of $512 \times 512$ in height and width (with only depth $D$ varying), and the feature size of $16\times16\times16 \times 768$ keeps the model within our GPU memory (48 GB VRAM).

\section{3D Scanpath Metrics}
\label{sec:metrics}


Due to the complexity of 3D scanpath metrics, we describe them at a high level and point out the modifications from 2D to 3D. For detailed implementation, we encourage readers to examine the source code directly:
\begin{itemize}
    \item \verb|visual_attention_metrics.py|: Contains implementations of:
          \begin{itemize}
              \item Saliency metrics Linear Correlation Coefficient (CC), Normalized Scanpath Saliency (NSS), Kullback-Leibler divergence (KLDiv).
              \item String-edit-distance (SED) metric.
          \end{itemize}
    \item \verb|scanmatch3d.py|: Contains the 3D-adapted version of ScanMatch.
    \item \verb|multimatch_3dgaze.py|: Contains the 3D-adapted version of MultiMatch.
\end{itemize}

\subsection{Saliency Metrics}
All three saliency metrics, CC (Correlation Coefficient), NSS (Normalized Scanpath Saliency), and KLDiv (Kullback-Leibler Divergence), are based on heatmaps and can be used directly without modification.
The CC metric is defined as:
\begin{align}
       & \hat{S} = \frac{S - \mu_S}{\sigma_S} \notag \\
       & \hat{G} = \frac{G - \mu_G}{\sigma_G} \notag \\
       & CC(S,G) = \frac{\sum_{i,j,k} \hat{S}_{ijk}\hat{G}_{ijk}}{\sqrt{\sum_{i,j,k} \hat{S}_{ijk}^2 \sum_{i,j,k} \hat{G}_{ijk}^2}}
\end{align}
where $S\in [0,1]^{H \times W \times D}$ is the saliency map, $G \in \{0,1\}^{H \times W \times D}$ is the ground truth fixation map, and $\mu$ and $\sigma$ are mean and standard deviation. Given the fixation sequence $\{(x_l,y_l,z_l)\}_{l=1}^N$, where $N$ is the fixation length, the ground truth map is defined as:
\begin{equation}
    G_{ijk} = \begin{cases}
1 & \text{if } (i,j,k) \in \{(x_l,y_l,z_l)\}_{l=1}^N \\ 
0 & \text{otherwise}
\end{cases}
\end{equation}
Higher CC scores indicate better matching between sequences, with an upper bound of 1.0.

The NSS metric is defined as: 
\begin{equation}
    \begin{aligned}
        &   \tilde{S} = \begin{cases}
            \frac{S}{\max(S)} & \text{if } \max(S) \neq 0 \\
            S & \text{otherwise}
        \end{cases} \\
        &  \hat{S} = \begin{cases}
            \frac{\tilde{S} - \mu_{\tilde{S}}}{\sigma_{\tilde{S}}} & \text{if } \sigma_{\tilde{S}} \neq 0 \\
            \tilde{S} & \text{otherwise}
        \end{cases} \\
        &   NSS(\hat{S},F) = 
            \frac{1}{N} \sum_{i,j,k} \hat{S}_{ijk}G_{ijk}
    \end{aligned}
    \end{equation}
where $\hat{S}$ is the normalized saliency map. Higher NSS scores indicate better matching between sequences, with an upper bound that depends on the ground truth.

The KLDiv metric is defined as: 
\begin{equation}
    \begin{aligned}
        & \tilde{S}_{ijk} = \frac{S_{ijk}}{\sum_{i,j,k} S_{ijk}} \\
        & \tilde{G}_{ijk} = \frac{G_{ijk}}{\sum_{i,j,k} G_{ijk}} \\
        & KLDiv(S,G) = \sum_{i,j,k} \tilde{G}_{ijk} \log \left(\epsilon + \frac{\tilde{G}_{ijk}}{\tilde{S}_{ijk} + \epsilon}\right)
    \end{aligned}
    \end{equation}
where $\epsilon = 2.2204 \times 10^{-16}$ is a small constant to prevent divide-by-zero and log-zero. Lower KLDiv scores indicate better matching between sequences, with a lower bound of 0.0.

\subsection{String-edit-distance}
String-edit-distance (SED) has two main steps:
\begin{enumerate}
    \item Converts gaze sequences into strings:
        \begin{enumerate}[label=\alph*)]
            \item Dividing the 3D volume into discrete cells. In the original 2D version, this step divides the image into patches.
            \item Assigning unique characters to each cell.
            \item Mapping fixation points to these characters in sequence.
        \end{enumerate}
    \item Compares two sequences using Levenshtein distance by counting minimum number of operations (insertions, deletions, substitutions).
\end{enumerate}
In our 3D adapted SED, we change the step 1.a from 2D into 3D, the other steps are left as is.
Lower SED scores indicate better matching between sequences, with a lower bound of 0.0.

\subsection{ScanMatch}
Calculating ScanMatch (SM) score between predicted and ground truth fixations consists of 3 main steps:
\begin{enumerate}
    \item Convert fixation sequences into letter strings. This step is similar to the first step of SED. In addition, when considering duration (ScanMatch w/ Dur.), each character is repeated n times, where n is the duration in milliseconds. This repetition is not performed when duration is not considered (ScanMatch w/o Dur.).
    \item Create a substitution matrix with scores for all possible letter pairs. The original score function uses 2D Euclidean distance, which we extend to 3D Euclidean distance.
    \item Sequence comparison:
    \begin{enumerate}[label=\alph*)]
        \item Create comparison matrix:
            \begin{itemize}
                \item Columns: letters from first sequence
                \item Rows: letters from second sequence
                \item Cell values: costs from substitution matrix
            \end{itemize}
        \item Apply Needleman-Wunsch algorithm to find optimal alignment path
        \item Calculate normalized similarity score (0-1 scale)
    \end{enumerate}  
\end{enumerate}
We adapt to 3D by modifying both step 1 and the substitution matrix calculation at step 2, replacing 2D Euclidean distance with 3D Euclidean distance.
Higher SM scores indicate better matching between sequences, with an upper bound of 1.0. 

\subsection{MultiMatch}
Different from ScanMatch and SED, MultiMatch (MM) measures scanpath similarity regarding shape, direction, length, position, and duration. Higher MM scores indicate better sequence matching, with an upper bound of 1.0 for all aspects.
Given the predicted fixations $\{(\hat{x}_l,\hat{y}_l,\hat{z}_l,\hat{t}_l)\}_{l=1}^N$ and ground truth fixations $\{(x_l,y_l,z_l,t_l)\}_{l=1}^N$, we calculate MultiMatch scores with 3 main steps:
\begin{enumerate}
    \item Temporal alignment:
        \begin{enumerate}[label=\alph*)]
            \item Calculate how similar each element $i$ in one scanpath is compared to each element $j$ in the other scanpath based on a similarity metric. Collect all pairs $(i,j)$ to create a similarity matrix $M(i,j)$ between elements.
            \item From $M(i,j)$, build adjacency matrix $A$ with connection weights like a graph.
            \item Find the shortest path from $i$ to $j$ using Dijkstra's algorithm.
            \item Align scanpaths along shortest path.
        \end{enumerate}
    
    \item For every align pair of fixation $(i,j)$, we compute similarity across five dimensions:
        \begin{enumerate}[label=\alph*)]
            \item \textbf{Vector (shape):} shape difference between fixation vectors $(\hat{x}_i,\hat{y}_i,\hat{z}_i,\hat{t}_i)-(x_j,y_j,z_j,t_j)$.
            \item \textbf{Direction:} difference in direction (angle) between fixation vectors. We measure angles using spherical coordinates in 3D, analogous to the original authors' use of polar coordinates in 2D.
            \item \textbf{Length:} difference in amplitude (length) between fixation vectors $\left|(\hat{x}_i,\hat{y}_i,\hat{z}_i,\hat{t}_i)-(x_j,y_j,z_j,t_j)\right|$,
            \item \textbf{Position:}: 3D Euclidean distance between fixations.
            \item \textbf{Duration:}: difference in duration between fixations.
        \end{enumerate}
    
    \item Score normalization:
        \begin{enumerate}[label=\alph*)]
            \item Vector, Length, and Position scores are normalized by volume diagonal.
            \item Direction is normalized by $\pi$.
            \item Duration is normalized by maximum duration.
        \end{enumerate}
\end{enumerate}

\section{3D Scanpath Prediction Baselines}
\label{sec:add-baseline}
\subsection{PathGAN}
\begin{figure}[h]
    \centering
    \includegraphics[width=0.8\linewidth]{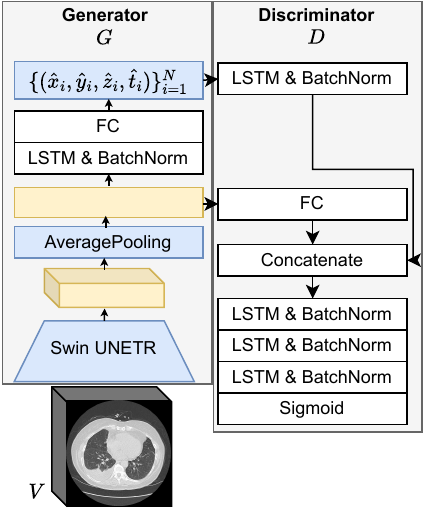}
    \caption{Our adapted version of PathGAN for CT scanpath prediction maintains the core architecture while altering components: the visual encoder module to Swin UNETR, Average Pooling to 3D, and predicted fixations (highlighted in blue).}
    \label{fig:3dpathgan}
\end{figure}
Similar to original PathGAN~\cite{Assens2018pathgan}, our CT-adapted PathGAN architecture has two major components: a Discriminator $D$ and a Generator $G$. The Generator takes a CT volume $V$ as input and produces a fixation sequence \(G(V)=\{(\hat{x}_i, \hat{y}_i, \hat{z}_i, \hat{t}_i)\}_{i=1}^N\). The Discriminator aims to assign low scores to $N$ predicted fixations \((\hat{x}_i, \hat{y}_i, \hat{z}_i, \hat{t}_i)\}_{i=1}^N\) and high scores to $N_{gt}$ ground truth fixations \(gt=\{(x_i, y_i, z_i, t_i)\}_{i=1}^{N_{gt}}\), taking both the fixation sequence and CT volume features as input.
The PathGAN architecture is illustrated in \cref{fig:3dpathgan}. Note that we share a frozen Swin UNETR module between $G$ and $D$ during training while optimizing other modules.

For the loss functions, we maintain PathGAN's default implementation using two main components: conditional GAN loss and $L^2$ loss between ground truth and predicted fixations. Specifically, the conditional GAN loss is defined as:
\begin{equation}
    \resizebox{0.9\linewidth}{!}{
        $\mathcal{L}_{\mathrm{cGAN}} = \mathbb{E}_{V, gt}[\log D(V, gt)]+\mathbb{E}_{V}[\log (1-D(V, G(V))],$
    }
\end{equation}
The L2 loss is defined as:
\begin{equation}
    \mathcal{L}_{L^2}=\mathbb{E}_{V, gt}\left[\|gt-G(V)\|^2\right]
\end{equation}
The final formulation of the loss function for the generator during adversarial training is:
\begin{equation}
    \mathcal{L}=\mathcal{L}_{\mathrm{cGAN}}+\alpha \mathcal{L}_{L^2}
\end{equation}
Following the original PathGAN implementation, we set the hyperparameter $\alpha=0.05$.

\subsection{HAT}

\begin{figure}[t]
    \centering
    \includegraphics[width=\linewidth]{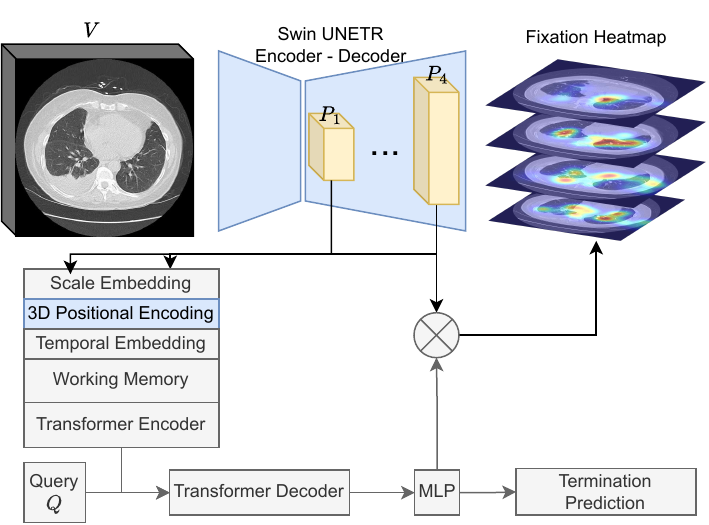}
    \caption{Our adapted version of HAT for CT scanpath prediction maintains most of the original architecture while modifying key components: the Visual Encoder (Swin UNETR), 3D Position Encoding, and prediction output (highlighted in blue). $\otimes$ is matrix multiplication.}
    \label{fig:3dhat}
\end{figure}
The detailed adapted HAT~\cite{yang2024hat} architecture is shown in \cref{fig:3dhat}.
Similar to HAT's FPN visual encoder~\cite{lin2017fpn} that generates features at both bottleneck and high-resolution levels, we extract features from two Swin UNETR layers: $P1 \in \mathbb{R}^{H/32 \times W/32 \times D/32 \times C}$ from the bottleneck layer as low resolution feature and $P4 \in \mathbb{R}^{H/4 \times W/4 \times D/4 \times C_4}$ from the $4^{th}$ layer of Swin UNETR's decoder as high resolution feature. We extend HAT's original loss to handle 3D fixation maps and use a single query as the class query. Note that \cref{fig:3dhat} shows one-step prediction because HAT predicts fixations step by step, with the working memory being updated after each fixation heatmap prediction~\cite{yang2024hat}.

Then, given predicted fixation heatmaps $\hat{Y} \in [0,1]^{H \times W \times D}$ and termination probabilities $\hat{\tau} \in [0,1]$, we compute the loss at each step $i$:
\begin{equation}
    \mathcal{L}=\frac{1}{N} \sum_{i=1}^N \mathcal{L}_{\text {fix }}\left(\hat{Y}_i, Y_i\right)+\mathcal{L}_{\text {term }}\left(\hat{\tau}_i, \tau_i\right)
\end{equation}
where $Y_i \in [0,1]^{H \times W \times D}$ represents the ground-truth 3D fixation heatmap, $\tau_i \in\{0,1\}$ is the termination label, and $N$ is the length of the fixation sequence. We generate $Y$ by applying a Gaussian kernel with sigma equal to 1 visual angle to the ground-truth fixation map. The fixation loss $\mathcal{L}_{\text {fix }}$ is a volumetric focal loss:
\begin{equation}
    \resizebox{0.9\linewidth}{!}{
    $\mathcal{L}_{\text {fix }}=\frac{-1}{H W D} \sum_{i,j,k}\left\{\begin{array}{cc}
            \left(1-\hat{Y}_{ijk}\right)^\alpha \log \left(\hat{Y}_{ijk}\right) & \text { if } Y_{ijk}=1 \\
            \left(1-Y_{ijk}\right)^\beta\left(\hat{Y}_{ijk}\right)^\alpha       &                        \\
            \log \left(1-\hat{Y}_{ijk}\right)                                   & \text { otherwise }
        \end{array}\right.$
    }
\end{equation}
with $\alpha=2$ and $\beta=4$. And the termination loss $\mathcal{L}_{\text {term }}$ is a binary cross entropy loss:
\begin{equation}
    \mathcal{L}_{\text {term }}=- \tau \log \left(\hat{\tau}_i\right)-(1-\tau) \log \left(1-\hat{\tau}_i\right)
\end{equation}

\subsection{Gazeformer}
\begin{figure}[t]
    \centering
    \includegraphics[width=\linewidth]{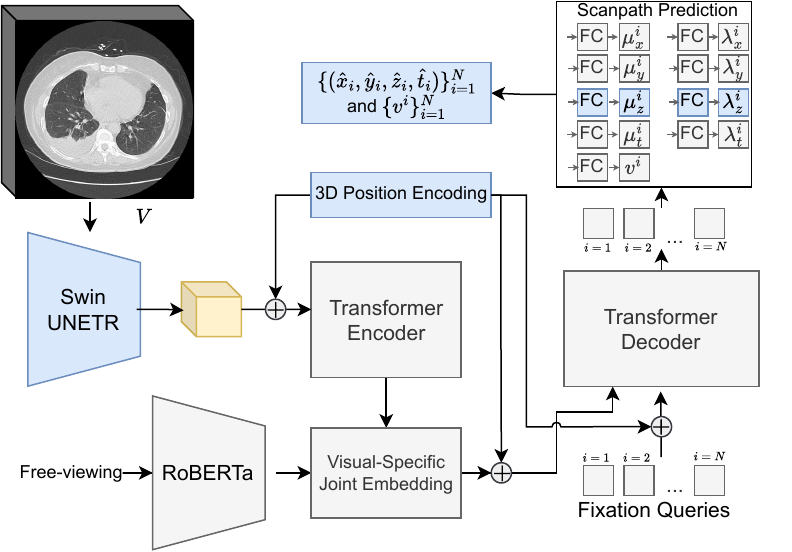}
    \caption{Our adapted version of Gazeformer for CT scanpath prediction maintains most of the original architecture while modifying three key components (highlighted in blue): replacing the visual encoder with Swin UNETR, incorporating 3D Position Encoding, and adding an extra prediction head for z-coordinate distribution in the Scanpath Prediction module. We use `freeview' as input for RoBERTa~\cite{liu2019roberta} and modify the Scanpath Prediction module to predict three separate branches for x, y, and z coordinates. }
    \label{fig:3dgazeformer}
\end{figure}
Our adapted version of Gazeformer for CT scanpath prediction is illustrated in \cref{fig:3dgazeformer}. Besides adapting the architecture to our task, we also extend the loss function to 3D.
The total loss function is defined as:
\begin{equation}
    \mathcal{L}=\left(\mathcal{L}_{xyzt}+\mathcal{L}_{val}\right)
\end{equation}
where the coordinate regression loss is:
\begin{equation}
    \resizebox{0.9\linewidth}{!}{$
        \mathcal{L}_{xyzt}=\frac{1}{N_{gt}} \sum_{i=1}^{N_{gt}}\left(\left|x_i-\hat{x}_i\right|+\left|y_i-\hat{y}_i\right|+\left|z_i-\hat{z}_i\right|+\left|t_i-\hat{t}_i\right|\right)
    $}
\end{equation}
and the validity prediction loss is:
\begin{equation}
    \mathcal{L}_{val}=-\frac{1}{N} \sum_{i=1}^{N}\left(\hat{v}_i \log v_i+\left(1-\hat{v}_i\right) \log \left(1-v_i\right)\right)
\end{equation}

Here, $\left\{\left(\hat{x}_i, \hat{y}_i, \hat{z}_i, \hat{t}_i\right)\right\}_{i=1}^{N}$ represents the predicted scanpath and $N$ is the maximum predicted scanpath length. $N_{gt}$ denotes the length of the ground truth scanpath $\left\{\left( x_i, y_i, z_i, t_i  \right)\right\}_{i=1}^{N_{gt}}$. The binary scalar $v_i$ indicates whether the $i^{th}$ token in the ground truth fixation is a valid fixation or padding, while $\hat{v}_i$ represents our model's predicted probability of token validity.

\subsection{GazeformerISP}
\begin{figure}[t]
    \centering
    \includegraphics[width=\linewidth]{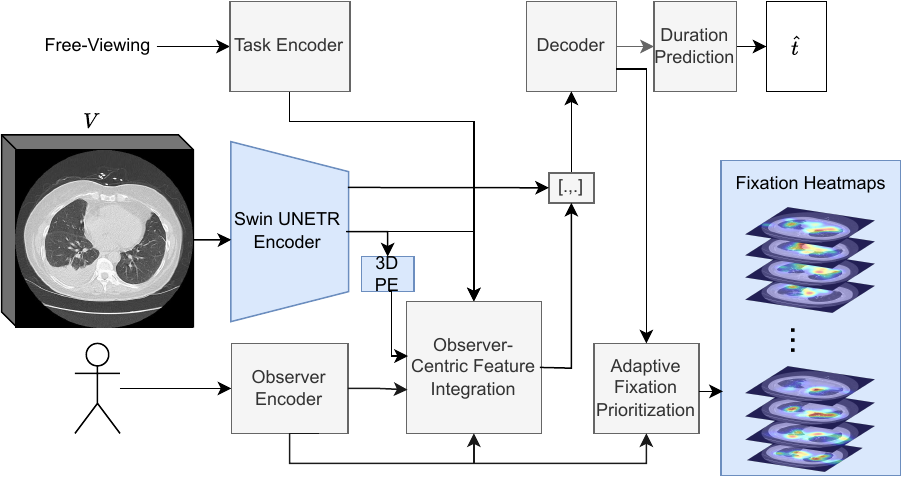}
    \caption{Our adapted version of GazeformerISP for CT scanpath prediction maintains most of the original architecture while modifying three key components (highlighted in blue): the input features, 3D Position Encoding, and prediction output. Here, [.,.] denotes the concatenation operator, and `3D PE' represents 3D Positional Encoding. Similar to HAT's freeview mode, we employ a single query embedding to predict fixations autoregressively. }
    \label{fig:3disp}
\end{figure}

Our adapted version of GazeformerISP for CT scanpath prediction replaces the original encoder with Swin UNETR encoder's bottleneck features, 3D Position Encoding, prediction output and extends the loss function to handle 3D fixation maps. The architecture is illustrated in \cref{fig:3disp}.
GazeformerISP predicts and computes loss on 3D fixation maps to represent 3D coordinates. The objective jointly optimizes the fixation map $\hat{Y}_i$ and duration $\hat{t}_i$:
\begin{equation}
    \mathcal{L}=-\sum_{i=1}^{N} Y_t \log \hat{Y}_t +  \sum_{t=1}^{N}\left|t_i-\hat{t}_i\right|
\end{equation}
where $N$ is the maximum length of fixations, $Y_t$ and $t_i$ represent the ground-truth fixation maps and fixation duration, respectively.
Finally, we train GazeformerISP with Self-Critical Sequence Training (SCST) set up using ScanMatch as reward, and the Consistency Divergence loss as originally described in \cite{chen2024isp,xianyu:2021:vqa}.

\begin{figure*}[t]
    \centering
    \includegraphics[width=0.9\linewidth]{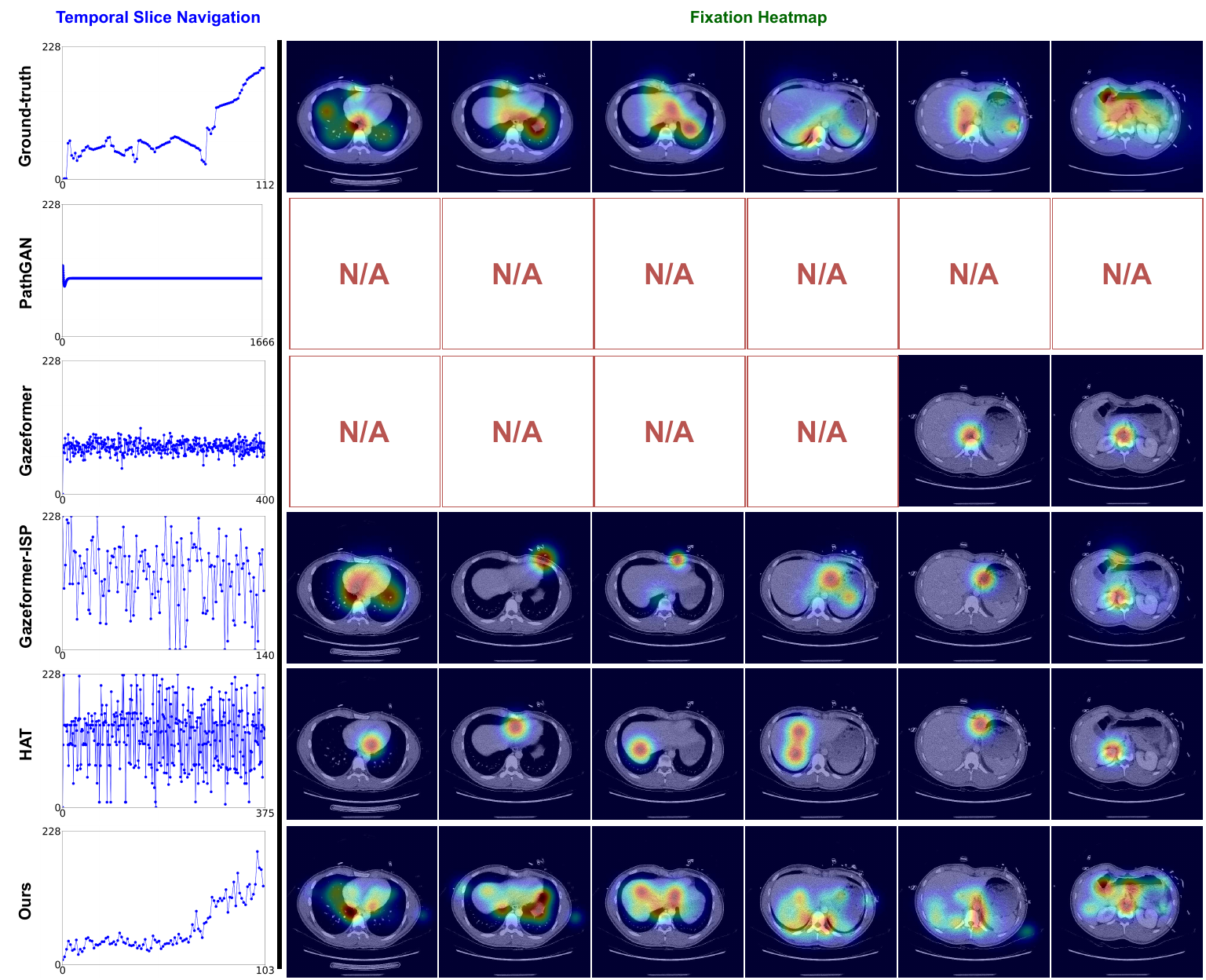}
    \caption{Additional qualitative results between \method and state-of-the-art scanpath prediction methods. N/A in a particular slice position (column) means that the corresponding model (row) fails to predict scanpath for that slice, thus no heatmap can be created. The heatmap images in the Fixation Heatmap column show the eye gaze fixation patterns across different CT slices. The left columns show Temporal Slice Navigation patterns, illustrating how scanpath traverses through different slices over time.}
    \label{fig:qual-result-sup}
\end{figure*}

\section{Additional Qualitative Results}
\label{sec:add-qual}

\cref{fig:qual-result-sup} presents an additional comparison of the temporal slice navigation and fixation heatmaps across multiple CT slices between \method with state-of-the-art scanpath prediction methods. \method outperforms others by capturing a balance between realism and variability, avoiding excessive noise or oversimplification in the temporal slice navigation comparison. Additionally, \method achieves more visually faithful heatmaps compared to the ground-truth, outperforming other approaches in detail and accuracy.
For baseline methods, we observe several limitations. Some methods produce no heatmaps (N/A) in certain positions, showing their inability to generate meaningful outputs. Similar to PathGAN, Gazeformer covers limited CT slices, indicating a constraint in handling 3D fixation tasks. Both GazeformerISP and HAT can produce heatmaps for most CT slices, however their temporal slice navigation appears noisy and inconsistent, deviating from the ground-truth pattern.
In conclusion, our method outperforms other approaches and mimics ground truth scanpath in both temporal slice navigation and fixation heatmap generation.
\begin{table}[t]
    \centering
    \caption{Ablation: Comparison of backbone architectures across scanpath metrics. Arrows ($\uparrow$/$\downarrow$) indicate whether higher or lower scores are better. Bold values indicate the best performance.}
    \resizebox{\linewidth}{!}{
    \begin{tabular}{c|c|c|c|c}
        \hline
        Visual Encoder & SM $\uparrow$   & MM $\uparrow$   & SED $\downarrow$ & KLDiv$\downarrow$ \\ \hline
        CT-ViT         & 0.1287          & 0.6934          & 193              & 3.665             \\
        Swin UNETR     & \textbf{0.1318} & \textbf{0.7002} & \textbf{174}     & \textbf{3.645} \\ \hline
    \end{tabular}
    }
    \label{tab:ablation-backbone}
\end{table}
\begin{figure*}[t]
    \centering
    \includegraphics[width=0.8\linewidth]{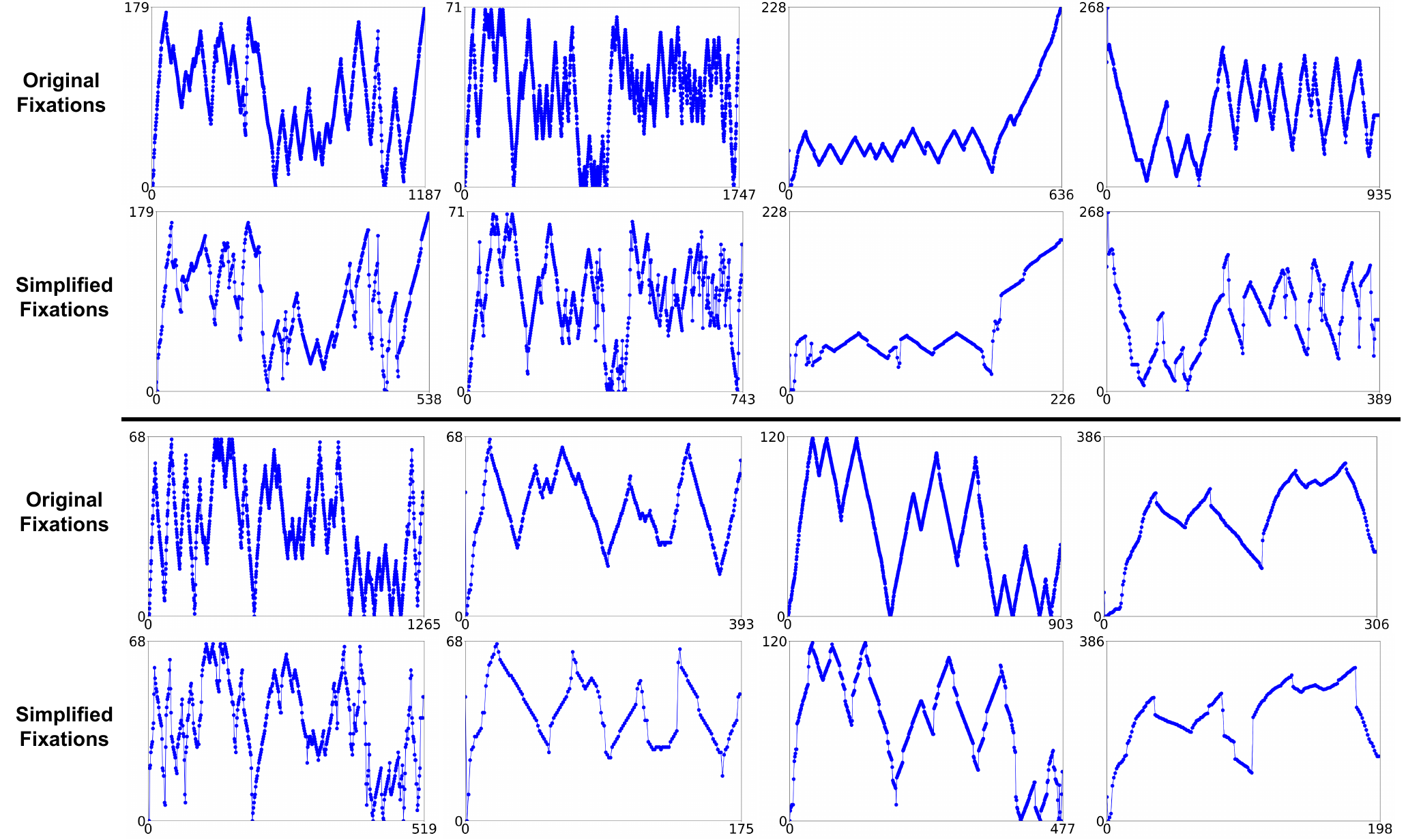}
    \caption{The effect of MM simplification algorithm on the z (slice) dimension.}
    \label{fig:data-sim-z}
\end{figure*}
\section{Comparison of CT Backbone Architectures}
\label{sec:compare-ct-backbone}

\cref{tab:ablation-backbone} demonstrates the comparative performance of CT-ViT and Swin UNETR backbones as \method's Visual Encoder across multiple scanpath similarity metrics: ScanMatch (SM), MultiMatch (MM), String-Edit Distance (SED), and Kullback-Leibler Divergence (KLDiv). We freeze both backbones during training. \cref{tab:ablation-backbone} shows that Swin UNETR outperforms CT-ViT across all metrics, achieving higher scores in pattern-based measures (SM: 0.1318 vs. 0.1287, MM: 0.7002 vs. 0.6934) and lower values in distance-based metrics (SED: 174 vs. 193, KLDiv: 3.645 vs. 3.665). Based on the empirical results, we adopt Swin UNETR as our Visual Encoder.

\begin{table}[t]
    \centering
    \caption{Fixation count comparison between the original gaze data and the simplified gaze data.}
    \resizebox{\linewidth}{!}{
\begin{tabular}{l|cc|c}
\hline
Version & Original & Simplified & Reduction (\%) \\ \hline
Number of Fixations & 2,234,920 & 954,311 & 57.3\% \\ \hline
\end{tabular}}
    \label{tab:fixation_counts}
\end{table}
\begin{table}[t]
    \centering
    \caption{MultiMatch similarity scores between the original gaze data and the simplified gaze data.}
    \resizebox{\linewidth}{!}{
\begin{tabular}{l|cccc|c}
\hline
Dimension & Vector & Direction & Length & Position & Average \\ \hline
Score & 0.993 & 0.853 & 0.989 & 0.944 & 0.945 \\ \hline
\end{tabular}
    }
    \label{tab:multimatch_scores}
\end{table}
\section{MultiMatch Simplification Analysis}
\label{sec:mm}
\begin{figure*}[t]
    \centering
    \includegraphics[width=0.8\linewidth]{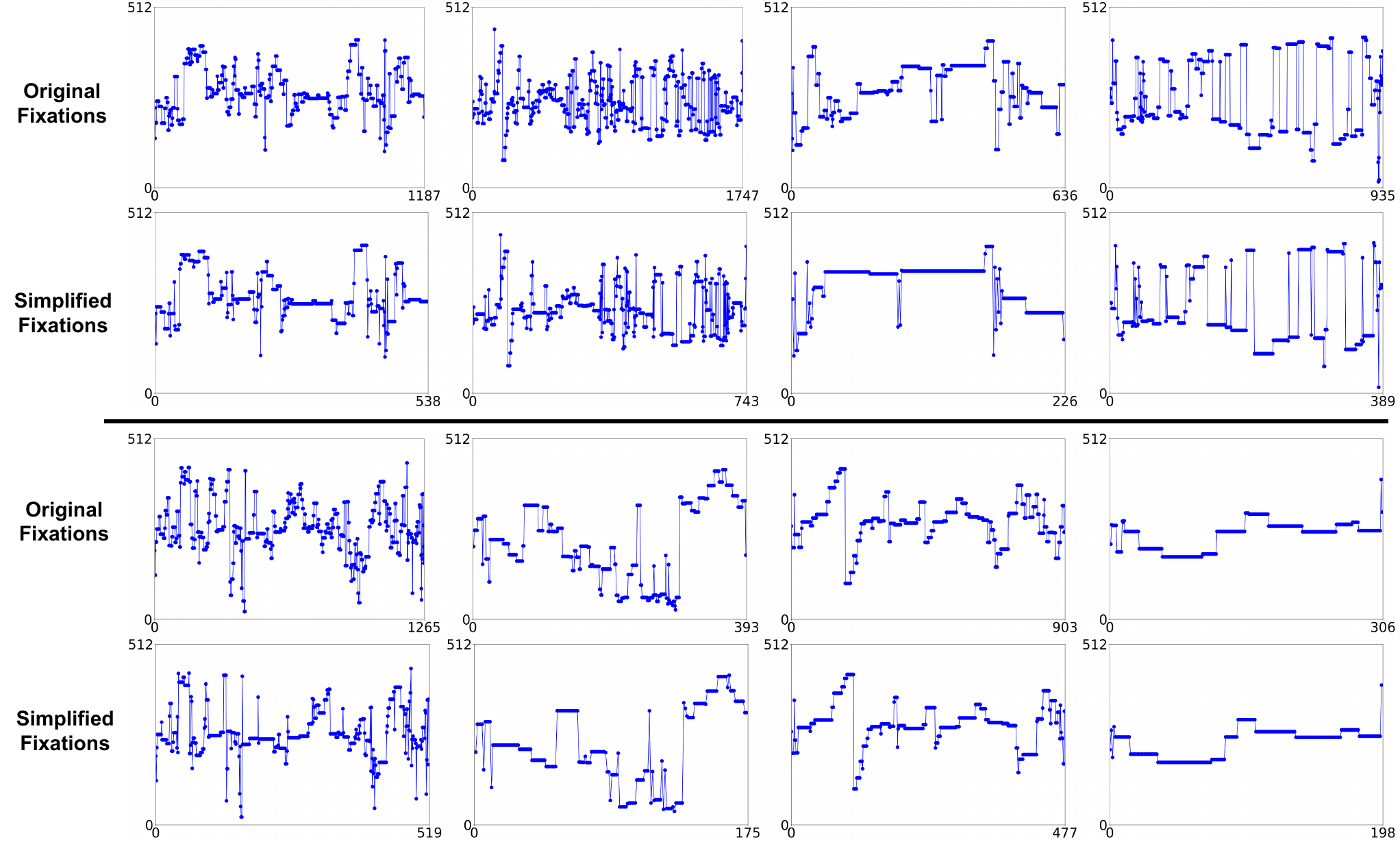}
    \caption{The effect of MM simplification algorithm on the x (width) dimension.}
    \label{fig:data-sim-x}
\end{figure*}
\begin{figure*}[t]
    \centering
    \includegraphics[width=0.8\linewidth]{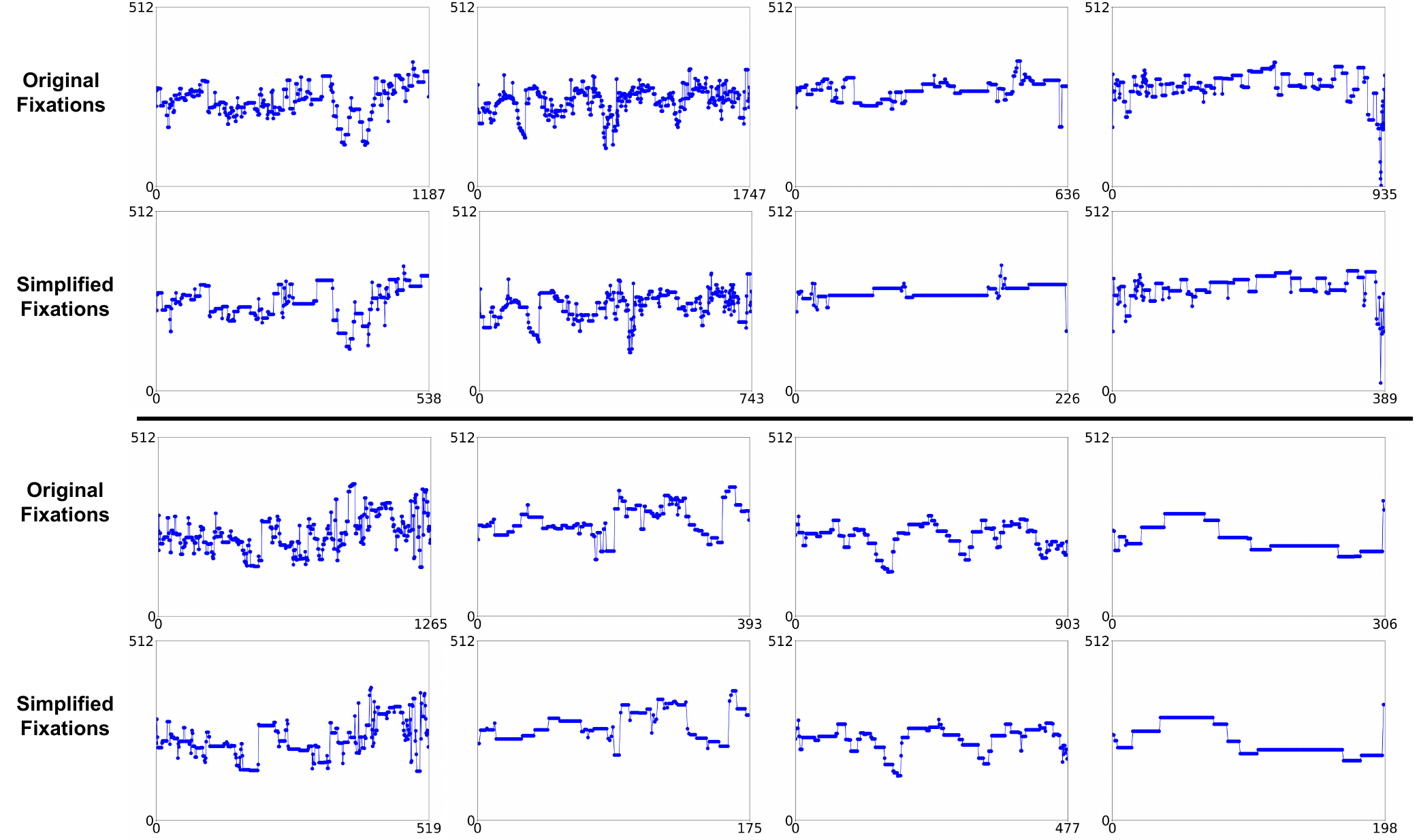}
    \caption{The effect of MM simplification algorithm on the y (height) dimension.}
    \label{fig:data-sim-y}
\end{figure*}

Due to the original gaze data containing scanpaths with numerous fixations that are dense and complex with sequences averaging 543 fixations and reaching up to 2,708 fixations per CT, we employ the simplification algorithm to make this data more manageable while preserving essential gaze patterns, from the MultiMatch toolbox~\cite{dewhurst2012depends} with default settings: an angular threshold of 45° and an amplitude threshold of 10\% of the volume resolution diagonal. This reduces sequences to an average of 222 fixations with a maximum of 1,507 fixations. Both original and simplified versions will be made available. 

To demonstrate the effectiveness of simplification, \cref{fig:data-sim-z,fig:data-sim-x,fig:data-sim-y} presents a comparative illustration between original and simplified scanpaths. While radiologists' eye movements on a single CT slice generally focus on the image center, movement along the slice dimension exhibits more complexity. \cref{fig:data-sim-z} reveals continuous and intricate radiologist navigation through depth. Nevertheless, the MM simplification approach introduces only minor changes to the movement landscape while preserving the overall pattern. This consistency in pattern preservation is also evident along the x-axis (\cref{fig:data-sim-x}) and y-axis (\cref{fig:data-sim-y}).
In summary, the line charts demonstrate that while significantly reducing the number of points (\cref{tab:fixation_counts}), the simplification process maintains the essential scanpath characteristics, as evidenced by minimal changes in MultiMatch similarity scores across all spatial dimensions (\cref{tab:multimatch_scores}).

\section{Discussion}
\label{sec:discussion}
Our contributions establish a foundation for volumetric scanpath modeling. \dataset benefits the research communities in several ways. First, it provides a benchmark specifically designed for 3D scanpath prediction in medical imaging, addressing limitations of 2D-focused datasets. Second, by capturing radiologists' visual attention patterns during diagnosis, it enables research on the relationship between visual search behavior and diagnostic reasoning, advancing explainable AI in heathcare. Furthermore,
\dataset enables several research directions: analyzing expert vs. novice radiologist gaze patterns, developing generalizable 3D attention models, creating radiology training protocols based on expert viewing patterns, and designing human-AI collaborative systems that leverage natural viewing behaviors.
\method demonstrates the feasibility of modeling expert visual behavior in CT interpretation. While our current implementation focuses on CT scans, the approach could extend to other domains requiring 3D visualization expertise, such as geological analysis or industrial CT inspection.
Future work should address current limitations, including expanding the dataset to encompass more diverse pathologies, developing more interpretable models that can explain predicted attention patterns, and evaluating the clinical impact of these systems on diagnostic accuracy and efficiency in real-world settings.

\clearpage

\end{document}